\documentclass[preprint,12pt]{elsarticle}

\usepackage[table,xcdraw]{xcolor}
\usepackage{graphicx}
\usepackage{multirow}
\usepackage{amsmath,amssymb,amsfonts}
\usepackage{amsthm}
\usepackage[title]{appendix}
\usepackage{textcomp}
\usepackage{booktabs}
\usepackage{array}
\usepackage{algorithmic}
\usepackage{icomma}
\usepackage[final]{microtype}
\usepackage[normalem]{ulem}
\usepackage{float}
\usepackage{rotating}
\usepackage{url}
\usepackage[hidelinks]{hyperref}

\biboptions{sort&compress}

\newcolumntype{L}[1]{>{\raggedright\arraybackslash}p{#1}}
\newcolumntype{C}[1]{>{\centering\arraybackslash}p{#1}}

\begin{document}

\begin{frontmatter}

\title{Rationally Enriched Chebyshev Trunk Bases for DeepONet Surrogates of High P\'eclet Entrance Transport}

\author[gatech]{Mingeun Choi}
\ead{mingeun.choi@gatech.edu}
\author[gatech]{Satish Kumar\corref{cor1}}
\ead{satish.kumar@me.gatech.edu}
\cortext[cor1]{Corresponding author.}

\affiliation[gatech]{organization={George W. Woodruff School of Mechanical Engineering, Georgia Institute of Technology},
                     city={Atlanta},
                     postcode={30332},
                     state={GA},
                     country={USA}}

\begin{abstract}

This study demonstrates a rationally enriched Chebyshev (REC) trunk for deep operator network (DeepONet) surrogate models of singularly perturbed and high-P\'eclet transport problems whose solution profiles are characterized by thin localized boundary or wall layers. The REC trunk combines Chebyshev polynomial dictionary elements with rational dictionary elements constructed using the adaptive Antoulas--Anderson (AAA) algorithm. Over five independent training runs, the resulting REC-trunk DeepONet is evaluated against a vanilla DeepONet and a Chebyshev-trunk DeepONet whose prescribed dictionary consists only of Chebyshev polynomials across three problems whose singular perturbation parameters are diffusion-to-advection ratios: a singularly perturbed scalar boundary-value problem (BVP), the thermal entrance problem with a prescribed wall temperature, and the concentration entrance problem with an absorbing wall. Across the held-out test profiles, the REC-trunk DeepONet improves over the vanilla DeepONet and remains comparable to the Chebyshev-trunk DeepONet in predicting the scalar profile, with its clearest advantage over the Chebyshev-trunk DeepONet appearing when the perturbation parameter lies between $1.00\times10^{-4}$ and $1.78\times10^{-4}$, where it reduces the profile-error metrics by up to $19.5\,\%$ relative to the Chebyshev-trunk DeepONet. In predicting the wall-normal temperature and concentration profiles, the REC-trunk DeepONet reduces the profile-error metrics by up to $60.2\,\%$ and $32.2\,\%$ relative to the vanilla and Chebyshev-trunk DeepONets, respectively, while suppressing artificial near-wall oscillations as the P\'eclet or mass-transfer P\'eclet number ranges from $10^{2}$ to $10^{4}$.

\end{abstract}


\end{frontmatter}

\section{Introduction}
\label{sec:introduction}

Transport problems are formulated as partial differential equations (PDEs), in which advection, diffusion, and reaction can interact across disparate spatial and temporal scales \cite{courant2008methods}. When diffusion is weak relative to advection, the associated boundary-value problem (BVP) can become singularly perturbed because the small perturbation parameter (or diffusion-to-advection ratio) multiplies the highest spatial derivative. Such problems often develop boundary or interior layers whose widths are far smaller than the domain length \cite{roos2008robust}. In resolving these localized layers, standard numerical schemes suffer from spurious oscillations or require a prohibitively dense grid \cite{roos2008robust}. Thus, accurate approximation of solution profiles requires techniques such as fitted finite differences, stabilized finite element methods, and layer-adapted meshes \cite{roos2008robust}, while repeated layer-resolving simulations across parameter ranges remain computationally expensive. To reduce this burden, recent machine-learning (ML) studies have developed physics-informed neural-network (PINN)-based models for singularly perturbed problems by incorporating asymptotic decompositions, stretched variables, or parameter continuation \cite{arzani2023theory,zhang2024multi,cao2023physics}. However, these models rely on problem-specific structures chosen from the differential equation, the small parameter, or the layer behavior of the target problem.

Deep operator networks (DeepONets) have emerged as alternative surrogates for singularly perturbed problems because they are designed to learn operators mapping sampled input functions and problem parameters to solution profiles through a branch--trunk representation \cite{lu2021learning,wang2021learning}. A pioneering application showed that a DeepONet can approximate sharp-gradient solutions by evaluating the training loss at layer-adapted Shishkin points, which depend on the perturbation parameter, rather than by modifying the network representation itself \cite{du2024approximation}. Subsequent studies then incorporated such layer-resolving priors within the operator-learning representation. Prandtl--Van Dyke DeepONet (PVD-ONet) decomposes the solution into outer, inner, and matching components and assigns these components to multiple DeepONet modules organized by Prandtl and Van Dyke matching \cite{sun2025pvd}. The enriched finite element operator network (eFEONet) augments the finite-element Galerkin ansatz with singular-perturbation corrector functions. The learned coefficients include both nodal finite-element coefficients and coefficients of the added layer-corrector basis \cite{lee2025efeo}. Physics-informed adaptive-scale DeepONet (PAS-Net) augments the trunk input with prescribed or learnable locally rescaled coordinates centered at reference points, thereby changing the coordinate representation seen by the trunk \cite{mou2025pas}. However, these approaches remain problem-specific, since the solution decomposition, finite element enrichment, or coordinate rescaling must be chosen for the particular layer being represented.

Parallel efforts have sought to reduce this dependence by expressing the output through prescribed or data-derived bases and their coefficients. Proper orthogonal decomposition DeepONet (POD-DeepONet) computes POD modes from the training outputs and uses the resulting modes as the trunk, leaving the branch network to predict modal coefficients \cite{lu2022fair}. Spectral coefficient learning via operator network (SCLON) predicts coefficients in orthogonal expansions such as Fourier or Legendre bases and applies this coefficient-space formulation to parametric PDEs ranging from singularly perturbed convection-diffusion equations to Navier--Stokes flows \cite{choi2023spectral}. The orthogonal polynomial neural operator (OPNO) builds neural operators around orthogonal-polynomial representations on bounded domains and treats Dirichlet, Neumann, and Robin boundary conditions within that polynomial framework \cite{liu2024render}. Spectral-embedded DeepONet (SEDONet) converts raw coordinate trunk inputs into values of Chebyshev polynomials before a trainable trunk network, improving bounded-domain DeepONet approximation for sharp gradients, boundary layers, and nonperiodic structures \cite{abid2025sedonet}. Nevertheless, these spectral and polynomial bases are global representations over the entire domain, and, thus, a thin localized layer can require many degrees of freedom unless the basis is enriched with dictionary elements adapted to the inner scale.

This inner-scale representation issue extends beyond idealized scalar BVPs because engineering transport often contains analogous high-P\'eclet convection-diffusion layer structure. In biomedical engineering, surface-based biosensors and related biofluidic capture systems involve convective delivery, diffusion, and surface binding near reactive walls \cite{squires2008making}. In chemical engineering, reactive microchannels and electrochemical chips generate wall-normal concentration fields governed by convective-diffusive delivery to reactive interfaces \cite{gervais2006mass,chevalier2021semianalytical}. In aerospace engineering, high-speed wall-bounded flows involve wall-normal heat and mass transfer in cooled and transpiration-cooled boundary layers \cite{sescu2019transpiration,hillcoat2025transpiration}, hypersonic boundary layers with strong near-wall thermal gradients \cite{xu2022hypersonic}, and reacting boundary layers with finite-rate wall chemistry \cite{passiatore2021finite,perakis2021recombination}. Across these examples, the common mathematical feature is a wall-normal temperature or concentration profile shaped by strong axial transport, transverse diffusion, and, in reactive cases, surface kinetics or wall absorption. Thus, entrance heat and mass transfer provide benchmark problems that retain the relevant wall-normal layer geometry under prescribed-wall, absorbing-wall, or Robin-type surface conditions \cite{shah1978laminar,haase2015graetz,popel1978mass,debarnot2018graetz,aquino2024equilibrium} while allowing accurate numerical reference profiles and controlled variation of P\'eclet number, axial location, inlet profile, and wall absorption strength. However, the use of inner-scale-enriched DeepONet trunk bases for such high-P\'eclet wall-normal profile reconstruction remains underexplored.

This paper introduces the rationally enriched Chebyshev (REC) trunk as a prescribed output-coordinate dictionary for DeepONet that combines Chebyshev polynomial dictionary elements with rational dictionary elements without adding trainable trunk parameters. This REC-trunk DeepONet is compared over five independent training runs with a vanilla DeepONet and a Chebyshev-trunk DeepONet, whose prescribed trunk dictionary consists only of Chebyshev polynomial dictionary elements, across three bounded-domain benchmark problems whose solution profiles contain one-sided localized layers. The first problem is the singularly perturbed scalar BVP, serving as the simplest bounded-interval formulation with an exponentially thin outflow layer \cite{du2024approximation}. The second problem is the thermal entrance problem with a prescribed wall temperature, where the temperature profile develops under hydrodynamically developed laminar duct flow \cite{shah1978laminar,haase2015graetz}. The third problem is the concentration entrance problem with an absorbing wall, where the concentration profile develops under the same hydrodynamically developed laminar duct flow \cite{popel1978mass,debarnot2018graetz}. Results demonstrate that DeepONet with the REC trunk remains globally competitive on the singularly perturbed scalar BVP and delivers its clearest advantage on the thermal and concentration entrance problems, where wall-attached layers control the profile geometry, by answering the following four questions:
\begin{enumerate}
\item How does the REC trunk perform on the singularly perturbed scalar BVP in the small-perturbation-parameter regime?
\item How does the REC trunk perform on the thermal entrance problem in the high-P\'eclet regime, when the learned output is a wall-normal profile at a queried axial location?
\item How does the REC trunk perform on the concentration entrance problem in the high-P\'eclet regime, under Robin wall kinetics and across different Damk\"ohler numbers?
\item How reproducible are the gains obtained with the REC trunk across repeated training runs and previously unseen solution profiles?
\end{enumerate}

\section{Methods}\label{sec2}

\subsection{Benchmark problems and operator-learning formulation}\label{sec2a}

\begin{figure*}[!t]
\centerline{\includegraphics[width=1.00\textwidth]{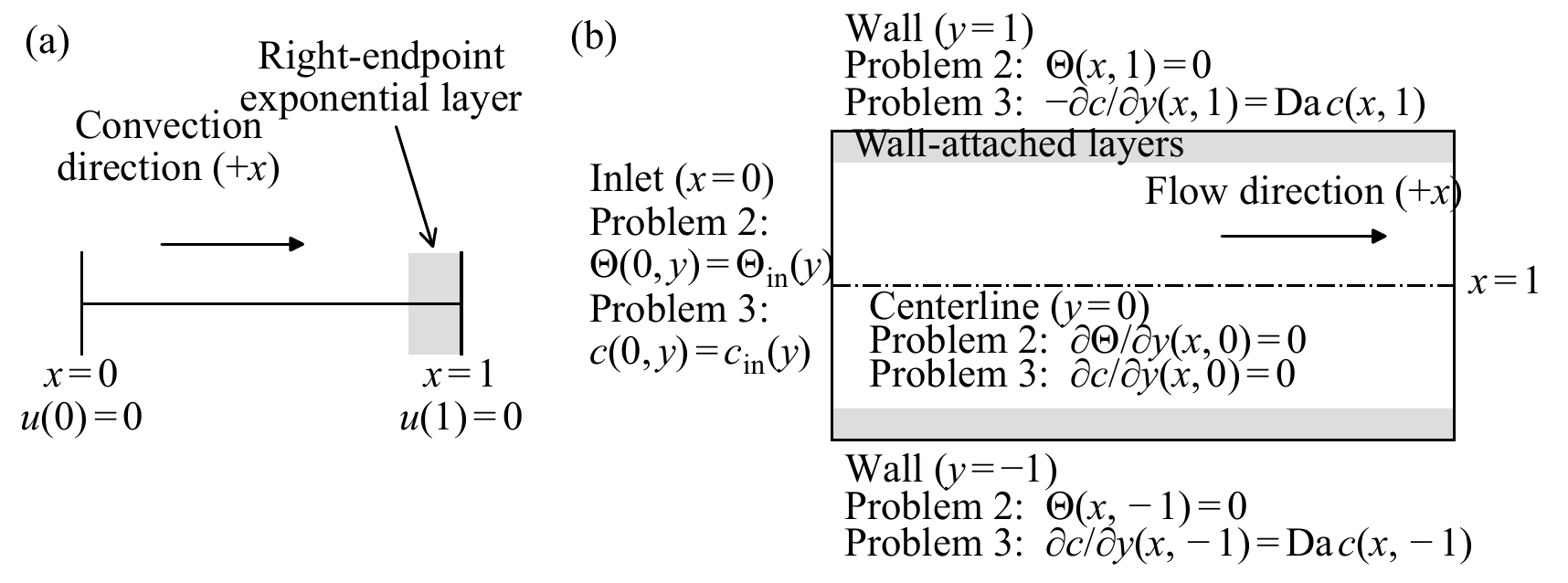}}
\caption{Computational domains and boundary conditions for (a) Problem~1 and (b) Problems~2 and~3. The full channel shown in (b) is the symmetric extension of the half-channel computational domain $y\in[0,1]$.}
\label{fig:computational_domains}
\end{figure*}

Figure~\ref{fig:computational_domains} illustrates the computational domains and boundary conditions for the three benchmark problems considered in this study. \textit{Problem 1} examines the singularly perturbed scalar BVP posed on the one-dimensional domain shown in Fig.~\ref{fig:computational_domains}(a) and defined by the equation \cite{du2024approximation}
\begin{equation}
-\varepsilon \frac{d^2 u}{dx^2}(x) + \frac{d u}{dx}(x) + u(x) = q(x), \qquad x \in (0,1),
\label{eq:problem1_bvp}
\end{equation}
with homogeneous Dirichlet boundary conditions
\begin{equation}
u(0)=u(1)=0.
\end{equation}

\noindent Here, $x$ denotes the dimensionless spatial coordinate on the bounded interval $(0,1)$, $u(x)$ denotes the dimensionless scalar solution field, $q(x)$ denotes the dimensionless randomized source term, and $\varepsilon$ denotes the singular perturbation parameter, given by the diffusion-to-advection ratio after the constant positive convective velocity is normalized to unity. The reaction coefficient is also fixed at unity. These choices allow the effects of $\varepsilon$ and $q(x)$ on the solution profiles to be examined without simultaneous changes in convection or reaction. Since $\varepsilon$ multiplies the highest-order diffusion term, decreasing its value weakens diffusion relative to convection, producing a thinner right-endpoint exponential layer \cite{roos2008robust}. The parameter range is sampled as $\varepsilon \in [10^{-4},10^{-2}]$, isolating the strongly perturbed regime targeted throughout this study.

\textit{Problem 2} addresses the thermal entrance problem with a prescribed wall temperature, known as the Graetz problem, posed on the two-dimensional domain shown in Fig.~\ref{fig:computational_domains}(b) and governed by the equation \cite{shah1978laminar,haase2015graetz}
\begin{equation}
w(y)\,\frac{\partial \Theta}{\partial x}(x,y) = \frac{1}{\mathrm{Pe}}\,\frac{\partial^2 \Theta}{\partial y^2}(x,y), \qquad (x,y)\in (0,1]\times(0,1),
\end{equation}
with a parabolic velocity profile for hydrodynamically fully developed flow
\begin{equation}
w(y)=1-y^2,
\end{equation}
centerline symmetry
\begin{equation}
\frac{\partial \Theta}{\partial y}(x,0)=0,
\end{equation}
a homogeneous Dirichlet condition at the wall
\begin{equation}
\Theta(x,1)=0.
\end{equation}
The inlet condition is
\begin{equation}
\Theta(0,y)=\Theta_{\mathrm{in}}(y).
\end{equation}

\noindent Here, $x$ denotes the dimensionless axial coordinate along the channel, $y$ denotes the dimensionless wall-normal coordinate across the channel half-width, $\Theta(x,y)$ denotes the dimensionless temperature difference relative to the prescribed wall temperature, and $w(y)$ denotes the prescribed dimensionless axial velocity profile. Because the boundary conditions are already defined, the surrogate directly maps the randomized inlet profile $\Theta_{\mathrm{in}}(y)$ to the wall-normal temperature profile $\Theta(x,y)$ evaluated at a queried axial location $x\in[0.05,1.0]$. The thermal P\'eclet number $\mathrm{Pe}=UL/\alpha$ measures axial advection relative to transverse thermal diffusion, where $U$ denotes the characteristic axial velocity, $L$ denotes the reference length scale, and $\alpha$ denotes the thermal diffusivity. Sampling $\mathrm{Pe}^{-1}\in[10^{-4},10^{-2}]$, which is the singular perturbation parameter analogous to $\varepsilon$ in Problem 1, isolates the high-P\'eclet regime characterized by increasingly thin wall-attached thermal layers.

\textit{Problem 3} investigates the concentration entrance problem with an absorbing wall, posed on the same domain used in Problem~2 and expressed by the equation \cite{popel1978mass}
\begin{equation}
w(y)\,\frac{\partial c}{\partial x}(x,y) = \frac{1}{\mathrm{Pe}_m}\,\frac{\partial^2 c}{\partial y^2}(x,y), \qquad (x,y)\in (0,1]\times(0,1),
\end{equation}
with the same parabolic velocity profile as in Problem~2
\begin{equation}
w(y)=1-y^2,
\end{equation}
the same centerline symmetry condition
\begin{equation}
\frac{\partial c}{\partial y}(x,0)=0,
\end{equation}
the inlet condition
\begin{equation}
c(0,y)=c_{\mathrm{in}}(y),
\end{equation}
and a Robin condition at the wall \cite{debarnot2018graetz}
\begin{equation}
-\frac{\partial c}{\partial y}(x,1)=\mathrm{Da}\,c(x,1).
\end{equation}

\noindent Here, $c(x,y)$ denotes the dimensionless concentration field, $c_{\mathrm{in}}(y)$ denotes the randomized inlet concentration profile, $\mathrm{Pe}_m$ denotes the mass-transfer P\'eclet number, and $\mathrm{Da}$ denotes the Damk\"ohler number, representing the dimensionless wall-absorption strength in a partially absorbing channel \cite{aquino2024equilibrium}. Under this nondimensionalization, both spatial coordinates are scaled by the channel half-width $L$. The dimensionless groups are $\mathrm{Pe}_m=UL/D$ and $\mathrm{Da}=kL/D$, where $U$ denotes the characteristic axial velocity, $D$ denotes the molecular diffusivity, and $k$ denotes a first-order surface uptake coefficient with units of velocity. Sampling $\mathrm{Pe}_m^{-1}\in[10^{-4},10^{-2}]$, analogous to $\mathrm{Pe}^{-1}$ in Problem 2, isolates the high-P\'eclet mass-transfer regime characterized by increasingly thin wall-attached concentration layers. To examine the effect of wall-absorption strength, Problem~3 is evaluated at three wall-absorption strengths at $\mathrm{Da}=0.1$, $1.0$, and $10.0$. Similar to the thermal Graetz model, the surrogate maps the randomized inlet condition to the wall-normal concentration profile $c(x,y)$ evaluated at a queried axial location $x\in[0.05,1.0]$.

\subsection{Operator learning formulation and baseline trunks}\label{sec2b}

For each benchmark problem, the learning task is to approximate a profile-valued operator
\begin{equation}
\mathcal{G}(\mathbf{s},\mu)=u(\cdot;\mathbf{s},\mu),
\label{deeponet1}
\end{equation}
where $\mathbf{s}$ denotes the sensor values of the problem-specific input function and $\mu$ denotes the problem parameters. In Problem~1, $\mathbf{s}$ contains samples of the source term $q(x)$, whereas in Problems~2 and~3 it contains samples of the inlet temperature profile $\Theta(0,y)$ and the inlet concentration profile $c(0,y)$, respectively. For each query coordinate $\xi$, the predicted profile value is written in the standard DeepONet form
\begin{equation}
\mathcal{G}(\mathbf{s},\mu)(\xi)
\approx
\widehat{u}(\xi;\mathbf{s},\mu)
=
\mathbf{b}(\mathbf{s},\mu)\cdot\boldsymbol{\phi}(\xi).
\label{deeponet2}
\end{equation}

\noindent Here, $\mathbf{b}(\mathbf{s},\mu)=[b_1(\mathbf{s},\mu),\ldots,b_p(\mathbf{s},\mu)]$ denotes the branch coefficient vector and $\boldsymbol{\phi}(\xi)=[\phi_1(\xi),\ldots,\phi_p(\xi)]$ denotes the trunk feature vector, whose $k$th components are $b_k(\mathbf{s},\mu)$ and $\phi_k(\xi)$, respectively. Specifically, $\xi$ represents the interval coordinate $x$ in Problem~1 and the wall-normal coordinate $y$ in Problems~2 and~3. The parameter vector $\mu$ collects the sample-dependent parameters that condition the operator output, corresponding to $\varepsilon$ in Problem~1, $(\mathrm{Pe}^{-1}, x)$ in Problem~2, and $(\mathrm{Pe}_m^{-1}, x)$ in Problem~3. For Problem~3, separate operator surrogates are constructed for fixed values of $\mathrm{Da}$ instead of including the Damk\"ohler number in $\mu$. For notational brevity, the dependence on $(\mathbf{s},\mu)$ is omitted below whenever the input profile and parameter value are fixed. 

To isolate the effect of trunk-basis design, the surrogate models compared within each benchmark use the same branch-network architecture, namely a multilayer perceptron (MLP) with Gaussian error linear unit (GELU) activations and Xavier initialization. This MLP comprises three hidden layers of width 256, and its output dimension is fixed to $p=129$ to match the dimension of each compared trunk. Before entering the branch network, the first component of the parameter vector $\mu$ is encoded through its base-10 logarithm, with any remaining components appended in raw form.

Within this shared framework, three distinct trunk designs are compared. The branch network provides the sample- and parameter-dependent coefficients, while the trunk representation defines the corresponding output approximation family. The first baseline model is the standard coordinate-trunk DeepONet, referred to as the \textit{Vanilla} model, with a trunk MLP consisting of three hidden layers of width $128$ and an output dimension of $p=129$ \cite{lu2021learning}. Unlike the Chebyshev and REC trunks introduced in the following, this baseline uses a learned trunk MLP that takes both the profile coordinate and the encoded parameter vector as inputs. Thus, for this baseline, Eq.~\eqref{deeponet2} is understood with the coordinate-only trunk vector $\boldsymbol{\phi}(\xi)$ replaced by a learned, parameter-conditioned trunk vector $\boldsymbol{\phi}^{\mathrm{van}}(\xi,\mu)$.

The second baseline model is the Chebyshev-trunk DeepONet, labeled as the \textit{Chebyshev} model. While prior approaches, such as SEDONet, process a Chebyshev feature vector through an additional trainable network \cite{abid2025sedonet}, the present baseline uses the Chebyshev polynomials directly as the trunk dictionary. This formulation isolates the effect of the prescribed trunk dictionary without introducing an auxiliary trainable transformation. For this baseline, the trunk vector $\boldsymbol{\phi}(\xi)$ in Eq.~\eqref{deeponet2} is the Chebyshev dictionary

\begin{equation}
\begin{split}
\boldsymbol{\phi}^{\mathrm{cheb}}(\xi)
=&
\big[\phi^{\mathrm{cheb}}_1(\xi),\ldots,\phi^{\mathrm{cheb}}_{129}(\xi)\big] \\
=&
\big[T_0(2\xi-1),\ldots,T_{128}(2\xi-1)\big],
\end{split}
\label{chebtrunk}
\end{equation}
where $T_j$ denotes the Chebyshev polynomial of the first kind of degree $j$.

\subsection{Proposed REC trunk}\label{sec2c}

The proposed third model is the REC-trunk DeepONet, denoted as the \textit{REC} model, motivated by rational discretizations for singularly perturbed BVPs, where layer-oriented rational mappings resolve thin layers with fewer global polynomial degrees \cite{wang2010rational}. Unlike the Chebyshev trunk in Eq.~\eqref{chebtrunk}, whose trunk dictionary consists entirely of Chebyshev polynomial elements, the REC trunk combines a low-degree outer Chebyshev subdictionary for the smooth outer field with an inner rational subdictionary for the localized layer. The rational dictionary elements are constructed by the adaptive Antoulas--Anderson (AAA) algorithm and remain unchanged during DeepONet training \cite{nakatsukasa2018aaa}.

The inner rational subdictionary is derived from a canonical exponentially decaying layer family on the unit interval,
\begin{equation}
\psi_{\delta}(\zeta)=\exp\!\left(-\frac{1-\zeta}{\delta}\right),
\qquad \zeta\in[0,1],\qquad \delta\in[10^{-4},10^{-2}],
\label{eq16}
\end{equation}
where $\zeta$ denotes the dictionary coordinate and $\delta$ denotes the prototype decay parameter, whose sampled range matches the numerical range of $\varepsilon$, $\mathrm{Pe}^{-1}$, and $\mathrm{Pe}_m^{-1}$ used in the three benchmark problems defined in Section~\ref{sec2a}. This family represents a one-sided localized layer attached to $\zeta=1$, corresponding to the right endpoint $x=1$ in Problem~1 and to the wall endpoint $y=1$ in Problems~2 and~3.

The construction proceeds by choosing $129-N_{\mathrm{out}}$ logarithmically spaced values $\delta_k\in[10^{-4},10^{-2}]$, one for each rational dictionary element. For each sampled value $\delta_k$, the profile $\psi_{\delta_k}$ is evaluated on a $1025$-point Chebyshev--Lobatto grid in $\zeta\in[0,1]$. The AAA algorithm approximates each sampled profile by a barycentric rational function with at most $12$ retained terms and tolerances of $10^{-8}$ for both approximation error and support-point matching checks. The resulting rational approximant is used as one rational dictionary element and is written as
\begin{equation}
r_k(\zeta)=
\frac{\sum_{j=1}^{m_k} \dfrac{w_{k,j} f_{k,j}}{\zeta-z_{k,j}}}
{\sum_{j=1}^{m_k} \dfrac{w_{k,j}}{\zeta-z_{k,j}}},
\end{equation}
where $k$ indexes the rational dictionary element associated with the sampled layer profile $\psi_{\delta_k}$, $z_{k,j}$ denotes the $j$th support point selected by AAA from the $1025$ Chebyshev--Lobatto points, $f_{k,j}=\psi_{\delta_k}(z_{k,j})$ denotes the sampled layer value at $z_{k,j}$, $w_{k,j}$ denotes the corresponding barycentric weight, and $m_k$ denotes the number of retained support points for the $k$th rational dictionary element \cite{nakatsukasa2018aaa}. The stored support points, sampled values, and barycentric weights define the rational dictionary elements used by the REC trunk.

When inserted into the DeepONet trunk, these rational dictionary elements are evaluated at the profile coordinate $\xi$. For the three benchmarks, $\xi=x$ in Problem~1 and $\xi=y$ in Problems~2 and~3, with $x,y\in[0,1]$. No additional coordinate map is introduced, and the same dictionary elements are evaluated with $\zeta=\xi$. Thus, the trunk dictionary $\boldsymbol{\phi}(\xi)$ in Eq.~\eqref{deeponet2} for REC trunk can be written as

\begin{equation}
\begin{split}
\boldsymbol{\phi}^{\mathrm{REC}}(\xi)
=&
\big[\phi^{\mathrm{REC}}_1(\xi),\ldots,\phi^{\mathrm{REC}}_{129}(\xi)\big] \\
=&
\big[T_0(2\xi-1),\ldots,T_{N_{\mathrm{out}}-1}(2\xi-1),r_1(\xi),\ldots,r_{129-N_{\mathrm{out}}}(\xi)\big],
\end{split}
\label{eq:rec_trunk_dictionary}
\end{equation}
where the first $N_{\mathrm{out}}$ Chebyshev dictionary elements form the outer Chebyshev subdictionary, and the remaining $(129-N_{\mathrm{out}})$ rational dictionary elements form the inner rational subdictionary, yielding a total trunk dimension of $129$. The main discussion in Section~\ref{sec3} focuses on the REC trunk with $N_{\mathrm{out}}=16$, while Section~\ref{sec3e} analyzes additional REC trunks with $N_{\mathrm{out}}=33$, $65$, and $97$ to examine how the outer-inner split affects the approximation behavior. Appendix~\ref{secA1} gives an idealized representation argument for this outer-inner split. The Chebyshev expansion of $\psi_{\delta}$ shows that decreasing $\delta$ shifts non-negligible spectral weight toward higher polynomial degrees, whereas the inner rational subdictionary provides coverage across the sampled range $\log_{10}\delta\in[-4,-2]$. Under the idealized decomposition in Eq.~\eqref{eqa18}, the REC trunk assigns the smooth outer field and the thin boundary-attached layer to separate parts of the trunk dictionary, rather than imposing an explicit asymptotic formula.

\subsection{Input sampling and reference profile generation}\label{sec2d}

The training, validation, and test datasets are constructed by pairing sampled problem inputs with their corresponding numerical reference profiles. In Problem~1, each input consists of a randomized source term and the perturbation parameter $\varepsilon$. In Problems~2 and~3, each input consists of a randomized inlet profile, the inverse-P\'eclet-type transport parameter, and the queried axial location $x$. The source term in Problem~1 and the inlet profiles in Problems~2 and~3 are sampled at fixed Chebyshev--Lobatto sensor nodes, which include the interval endpoints and cluster near the boundaries. The resulting sensor vector is supplied to the branch network together with the encoded parameter values, and the corresponding reference profile is evaluated at fixed Chebyshev--Lobatto output nodes. In all three benchmarks, the source or inlet function is sampled at $129$ Chebyshev--Lobatto sensor locations, which include the interval endpoints and cluster toward the boundaries. The target profile is evaluated at $257$ Chebyshev--Lobatto output locations, providing denser resolution near the endpoint or wall region where the localized layer develops.

The randomized input functions are constructed as finite smooth expansions rather than taken from external data. In Problem~1, the source term is formed from four sine modes and two localized Gaussian components,
\begin{equation}
q(x)=
\sum_{m=1}^{4} a_m^{(q)}\sin(m\pi x)
+
\sum_{j=1}^{2} \widetilde{a}_j^{(q)}
\exp\!\left[
-\frac{1}{2}
\left(
\frac{x-x_j^{(q)}}{\ell_j^{(q)}}
\right)^2
\right].
\end{equation}
The modal and Gaussian amplitudes are drawn independently according to
$a_m^{(q)},\widetilde{a}_j^{(q)}\sim\mathcal{N}(0,1)$, while the Gaussian centers and widths are sampled as
$x_j^{(q)}\sim\mathcal{U}(0,1)$ and
$\ell_j^{(q)}\sim\mathcal{U}(0.03,0.20)$.
The perturbation parameter is sampled by drawing
$\log_{10}\varepsilon$ uniformly from $[-4,-2]$.

For Problems~2 and~3, the inlet function is generated from a positive baseline, four cosine modes, and two localized Gaussian components. This common inlet representation is denoted by $g_{\mathrm{in}}(y)$, where $g_{\mathrm{in}}=\Theta_{\mathrm{in}}$ for Problem~2 and $g_{\mathrm{in}}=c_{\mathrm{in}}$ for Problem~3. Before imposing the lower bound, the raw inlet function is
\begin{equation}
g_{\mathrm{raw}}(y)=
1+
\sum_{m=1}^{4} a_m^{(\mathrm{in})}\cos((m-1)\pi y)
+
\sum_{j=1}^{2} \widetilde{a}_j^{(\mathrm{in})}
\exp\!\left[
-\frac{1}{2}
\left(
\frac{y-y_j^{(\mathrm{in})}}{\ell_j^{(\mathrm{in})}}
\right)^2
\right].
\end{equation}
The cosine amplitudes are drawn according to
$a_m^{(\mathrm{in})}\sim\mathcal{N}(0,0.18^2)$, the Gaussian amplitudes as
$\widetilde{a}_j^{(\mathrm{in})}\sim\mathcal{N}(0,0.12^2)$, the Gaussian centers as
$y_j^{(\mathrm{in})}\sim\mathcal{U}(0,1)$, and the Gaussian widths as
$\ell_j^{(\mathrm{in})}\sim\mathcal{U}(0.05,0.18)$.
To exclude sign-changing inlet temperature differences in Problem~2, for which heating and cooling are equivalent up to a global sign change, and negative inlet concentrations in Problem~3, the inlet function used to generate each reference solution is defined as
\begin{equation}
g_{\mathrm{in}}(y)=\max\{g_{\mathrm{raw}}(y),10^{-3}\}.
\end{equation}
For Problems~2 and~3, the transport parameter is sampled by drawing
$\log_{10}\mathrm{Pe}^{-1}$ and $\log_{10}\mathrm{Pe}_m^{-1}$, respectively, uniformly from $[-4,-2]$. The queried axial location is sampled independently as $x\sim\mathcal{U}(0.05,1.0)$.

The target profiles are generated by problem-specific deterministic numerical solvers. Problem~1 is solved using an upwind finite-difference discretization on a piecewise-uniform Shishkin-type mesh whose transition point depends on $\varepsilon$ and which places a finer subgrid near the outflow boundary $x=1$ where the endpoint layer forms. For Problems~2 and~3, the entrance-transport equations are solved by marching in the axial coordinate from the randomized inlet profile to the queried location $x$. At each axial step, an implicit finite-difference discretization of the transverse diffusion operator on a uniform grid yields a tridiagonal linear system for the updated wall-normal profile. The centerline Neumann condition is imposed at $y=0$, while the wall condition at $y=1$ is imposed as a Dirichlet condition for Problem~2 and as a Robin condition for Problem~3.

For the convergence check, the same sampled source term and perturbation parameter in Problem~1 are solved on the $4096$-cell Shishkin mesh used for dataset generation and an $8192$-cell Shishkin mesh and compared after linear interpolation onto the same $257$ Chebyshev--Lobatto output locations. For Problems~2 and~3, the same inlet profile, inverse-P\'eclet-type parameter, and queried axial location are solved using the dataset resolution of $257$ wall-normal nodes with $400$ axial steps per unit length and using $513$ wall-normal nodes with $800$ axial steps per unit length, again compared on the same $257$ Chebyshev--Lobatto output locations. Across $32$ randomly sampled checks for each benchmark problem, the mean relative differences between the two numerical profiles are $4.60\times10^{-4}$ for Problem~1, $1.97\times10^{-4}$ for Problem~2, and $1.06\times10^{-4}$, $1.09\times10^{-4}$, and $1.43\times10^{-4}$ for Problem~3 at $\mathrm{Da}=0.1$, $1.0$, and $10.0$, respectively. The corresponding maximum relative differences are $9.30\times10^{-4}$, $6.64\times10^{-4}$, $4.30\times10^{-4}$, $4.53\times10^{-4}$, and $5.26\times10^{-4}$.

\subsection{Training and evaluation}\label{sec2e}

To quantify model accuracy, four profile-based error metrics are defined on the common discrete evaluation grid. Let $u$ denote the numerical reference profile at the queried condition, let $\widehat{u}$ denote the corresponding model prediction, and define the profile error as $e=\widehat{u}-u$. The relative discrete profile error is
\begin{equation}
E_{2}^{\mathrm{rel}}=\frac{\|e\|_{2}}{\|u\|_{2}},
\end{equation}
and the pointwise maximum error is
\begin{equation}
E_{\infty}=\|e\|_{\infty}.
\end{equation}
Here, $\|\cdot\|_{2}$ denotes the Euclidean norm of the profile values on the fixed output grid, and $\|\cdot\|_{\infty}$ denotes the maximum absolute value over the same grid.

The layer-focused maximum error measures the largest absolute discrepancy inside a problem-dependent layer strip $\Omega_s$,
\begin{equation}
E_{\max}^{\mathrm{layer}}=\|e\|_{\infty,\Omega_s}.
\end{equation}
For Problem~1, the layer strip is the right-endpoint boundary-layer region
\begin{equation}
\Omega_{\ell}=\{x_j:\ x_j \ge 1-C\,\varepsilon |\log \varepsilon|\},
\end{equation}
where $C$ denotes a layer-width factor. This definition selects the grid points in the interval $\left[1-C\varepsilon\left|\log\varepsilon\right|,1\right]$ adjacent to the outflow boundary $x=1$, where the outflow Dirichlet condition is satisfied through an endpoint layer under positive convection. In Problem 1, $\Omega_s=\Omega_{\ell}$ and $C=5$. For Problems~2 and~3, the layer strip is the near-wall region adjacent to the controlled or absorbing wall,
\begin{equation}
\Omega_{w}=\{y_j:\ y_j \ge 1-C\sqrt{\eta x}\},
\end{equation}
where $\eta=\mathrm{Pe}^{-1}$ in Problem~2, $\eta=\mathrm{Pe}_m^{-1}$ in Problem~3, and $C$ denotes a layer-width factor. In these problems, $\Omega_s=\Omega_w$ and $C=4$. The near-wall region in Problems~2 and~3 is used only to define a common near-wall error measure for comparing the three surrogates and is not intended as an asymptotic estimate of the physical Graetz-layer thickness.

The layer-aware error is defined problem-dependently to emphasize the localized layer,

\begin{equation}
E_{\mathrm{LA}}=
\begin{cases}
\left[
\displaystyle
\frac{\|e\|_{Q}^{2}
+\varepsilon \left\|\dfrac{d e}{d x}\right\|_{Q}^{2}
+\|e\|_{Q,\Omega_{\ell}}^{2}}
{\|u\|_{Q}^{2}
+\varepsilon \left\|\dfrac{d u}{d x}\right\|_{Q}^{2}
+\|u\|_{Q,\Omega_{\ell}}^{2}}
\right]^{1/2},
& \text{for Problem~1}, \\[2.0ex]
\left[
\displaystyle
\frac{\|e\|_{2,\Omega_{w}}^{2}}
{\|u\|_{2,\Omega_{w}}^{2}}
\right]^{1/2},
& \text{for Problems~2 and~3}.
\end{cases}
\end{equation}
Here, $\|\cdot\|_{Q}$ denotes the trapezoidal-quadrature norm on the fixed output grid. The restricted norms $\|\cdot\|_{Q,\Omega_{\ell}}$ and $\|\cdot\|_{2,\Omega_{w}}$ are evaluated over the corresponding layer strips. For Problem~1, the derivatives in $E_{\mathrm{LA}}$ are evaluated on the same nonuniform output grid using one-sided two-point differences at the endpoints and centered two-point differences at the interior nodes, with the same rule applied to the numerical reference and all model predictions.

All three models are trained with the same mean-squared-error (MSE) objective and the same training procedure. Each model uses $3000$ training samples and $500$ validation samples, and performance is evaluated on $500$ test profiles not used during training or validation. The Adam optimizer is used with learning rate $5\times10^{-4}$, batch size $64$, weight decay $10^{-6}$, gradient clipping at $5.0$, and an exponential learning-rate scheduler with decay factor $0.995$, for at most $250$ epochs. This is followed by a limited-memory Broyden--Fletcher--Goldfarb--Shanno (L-BFGS) quasi-Newton refinement with at most $80$ iterations and strong-Wolfe line search. The checkpoint used for testing is selected by the lowest validation value of $E_{\max}^{\mathrm{layer}}$ among the models obtained during Adam optimization and after L-BFGS refinement. Each trunk design is trained in five independent runs on the same data split, with model comparisons made seed by seed. Thus, the same $500$ test profiles are evaluated for each of the five trained instances, yielding $2500$ error evaluations per model and benchmark.

\section{Results and discussion}\label{sec3}

\subsection{Singularly perturbed scalar BVP}\label{sec3a}

\begin{figure*}[!t]
\centerline{\includegraphics[width=1.00\textwidth]{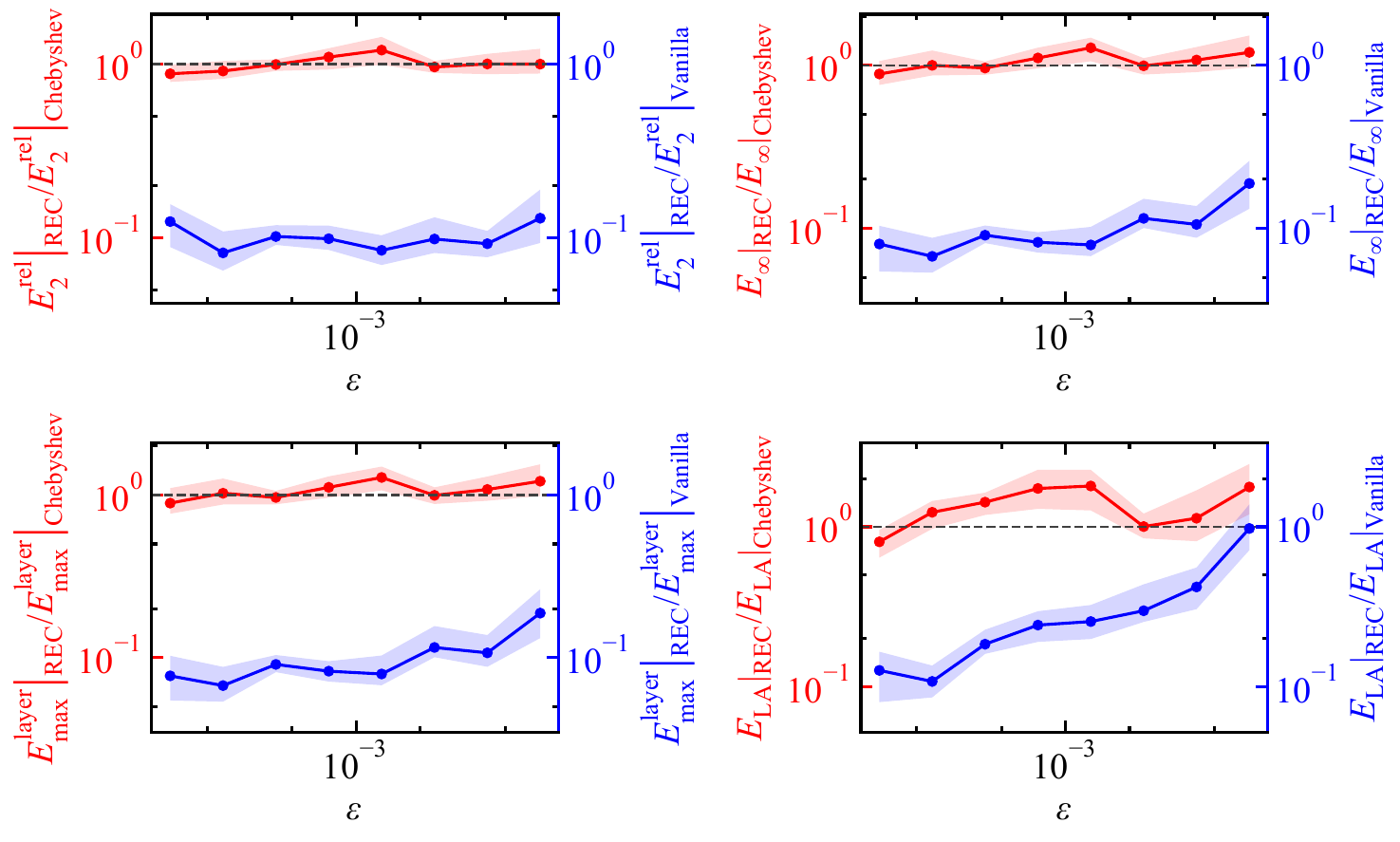}}
\caption{Ratios of $E_{2}^{\mathrm{rel}}$, $E_{\infty}$, $E_{\max}^{\mathrm{layer}}$, and $E_{\mathrm{LA}}$ over $\varepsilon\in[10^{-4},10^{-2}]$ for the singularly perturbed scalar BVP at $N_{\mathrm{out}}=16$.}
\label{fig:problem1_iqr}
\end{figure*}

Figure~\ref{fig:problem1_iqr} shows the ratios of the error from the \textit{REC} model to the error from the \textit{Chebyshev} model and to the error from the \textit{Vanilla} model over $\varepsilon\in[10^{-4},10^{-2}]$ for $E_{2}^{\mathrm{rel}}$, $E_{\infty}$, $E_{\max}^{\mathrm{layer}}$, and $E_{\mathrm{LA}}$. The interval $[10^{-4},10^{-2}]$ is divided into eight equal-width bins on a logarithmic scale, with nominal bin edges $1.00\times10^{-4}$, $1.78\times10^{-4}$, $3.16\times10^{-4}$, $5.62\times10^{-4}$, $1.00\times10^{-3}$, $1.78\times10^{-3}$, $3.16\times10^{-3}$, $5.62\times10^{-3}$, and $1.00\times10^{-2}$. For each test profile in a given bin, the error from the \textit{REC} model is divided by the error from the \textit{Chebyshev} model or the \textit{Vanilla} model for the same metric. The marker is placed at the geometric center of the bin and gives the median of these ratios. The shaded band gives the interquartile range (IQR), from the 25th to the 75th percentile. Relative to the \textit{Vanilla} model, the median ratios for the \textit{REC} model remain below one in all eight $\varepsilon$ bins and for all four metrics, with the largest median value reaching $0.977$ only for $E_{\mathrm{LA}}$ in the largest-$\varepsilon$ bin. Relative to the \textit{Chebyshev} model, the median ratios remain near unity across most $\varepsilon$ bins, whereas the smallest $\varepsilon$ bin gives ratios below unity for all four metrics, with median values $0.880$, $0.885$, $0.891$, and $0.805$ for $E_{2}^{\mathrm{rel}}$, $E_{\infty}$, $E_{\max}^{\mathrm{layer}}$, and $E_{\mathrm{LA}}$, respectively. These trends indicate that, for the singularly perturbed scalar BVP, the \textit{REC} model clearly separates from the \textit{Vanilla} model across the tested parameter range, while its advantage over the \textit{Chebyshev} model is concentrated where the perturbation parameter is smallest.

\begin{table*}[!t]
\caption{Comparison of the \textit{REC} model with the \textit{Vanilla} and \textit{Chebyshev} models for the singularly perturbed scalar BVP at $N_{\mathrm{out}}=16$.}
\centering
\begingroup
\footnotesize
\setlength{\tabcolsep}{1pt}
\renewcommand{\arraystretch}{1.12}
\begin{tabular}{L{\dimexpr0.330\textwidth-2\tabcolsep\relax}C{\dimexpr0.110\textwidth-2\tabcolsep\relax}C{\dimexpr0.180\textwidth-2\tabcolsep\relax}C{\dimexpr0.190\textwidth-2\tabcolsep\relax}C{\dimexpr0.190\textwidth-2\tabcolsep\relax}}
\toprule
Metric ratio & Full test set & First three parameter bins & Seeds with lower REC error & Profiles with lower REC error \\
\midrule
$\left.E_{2}^{\mathrm{rel}}\right|_{\mathrm{REC}}/\left.E_{2}^{\mathrm{rel}}\right|_{\mathrm{Vanilla}}$ & $0.107$ & $0.110$ & $5/5$ & $100\%$ \\
$\left.E_{\infty}\right|_{\mathrm{REC}}/\left.E_{\infty}\right|_{\mathrm{Vanilla}}$ & $0.0976$ & $0.0821$ & $5/5$ & $100\%$ \\
$\left.E_{\max}^{\mathrm{layer}}\right|_{\mathrm{REC}}/\left.E_{\max}^{\mathrm{layer}}\right|_{\mathrm{Vanilla}}$ & $0.0977$ & $0.0820$ & $5/5$ & $99.8\%$ \\
$\left.E_{\mathrm{LA}}\right|_{\mathrm{REC}}/\left.E_{\mathrm{LA}}\right|_{\mathrm{Vanilla}}$ & $0.335$ & $0.146$ & $5/5$ & $92.1\%$ \\
\cmidrule(lr){1-5}
$\left.E_{2}^{\mathrm{rel}}\right|_{\mathrm{REC}}/\left.E_{2}^{\mathrm{rel}}\right|_{\mathrm{Chebyshev}}$ & $0.962$ & $0.905$ & $5/5$ & $52.6\%$ \\
$\left.E_{\infty}\right|_{\mathrm{REC}}/\left.E_{\infty}\right|_{\mathrm{Chebyshev}}$ & $1.06$ & $0.970$ & $0/5$ & $43.9\%$ \\
$\left.E_{\max}^{\mathrm{layer}}\right|_{\mathrm{REC}}/\left.E_{\max}^{\mathrm{layer}}\right|_{\mathrm{Chebyshev}}$ & $1.06$ & $0.977$ & $0/5$ & $42.2\%$ \\
$\left.E_{\mathrm{LA}}\right|_{\mathrm{REC}}/\left.E_{\mathrm{LA}}\right|_{\mathrm{Chebyshev}}$ & $1.31$ & $1.11$ & $0/5$ & $32.8\%$ \\
\bottomrule
\end{tabular}
\endgroup
\label{tab:problem1_outer16_comparison}
\end{table*}

Table~\ref{tab:problem1_outer16_comparison} tabulates error ratios for the singularly perturbed scalar BVP at $N_{\mathrm{out}}=16$ for the \textit{REC} model against both the \textit{Vanilla} and \textit{Chebyshev} models. Here, the second column gives the ratio of mean errors over the five independent training runs, where each run error is averaged over the $500$ test profiles. The third column gives the ratio of mean errors over the five independent training runs, where each run error is averaged over the first three parameter-bin means using the same parameter bins as Fig.~\ref{fig:problem1_iqr}. The fourth column counts how many of the five independent training runs give lower error for the \textit{REC} model than for the denominator model. The fifth column gives the percentage of the $2500$ profile comparisons, from the five independent training runs and $500$ test profiles per run, in which the \textit{REC} model has lower error than the denominator model. Relative to the \textit{Vanilla} model, the second-column ratios are below $0.335$ for all four metrics, and the third-column ratios decrease further to at most $0.146$. The fourth and fifth columns show the same separation across all five independent training runs and at least $92.1\%$ of the $2500$ profile comparisons. Relative to the \textit{Chebyshev} model, the second column remains close to unity for $E_{2}^{\mathrm{rel}}$, $E_{\infty}$, and $E_{\max}^{\mathrm{layer}}$, while $E_{\mathrm{LA}}$ stays above unity. In the third column, however, $E_{2}^{\mathrm{rel}}$, $E_{\infty}$, and $E_{\max}^{\mathrm{layer}}$ decrease to $0.905$, $0.970$, and $0.977$, respectively. These results indicate that, for the singularly perturbed scalar BVP, the \textit{REC} model is most useful when the endpoint layer in $u(x)$ becomes thinnest, while the \textit{Chebyshev} model remains competitive when the full test set also includes larger perturbation parameters.

The REC reductions appear mainly in the first three parameter bins, while the full-test ratios relative to the \textit{Chebyshev} model remain near unity or above unity, consistent with the representation argument in Appendix~\ref{secA1}. The endpoint layer in Problem~1 belongs to the same exponential family used to construct the rational dictionary elements $r_k$ in Eq.~\eqref{eq:rec_trunk_dictionary}, while the same family also admits the exact Chebyshev expansion in Eq.~\eqref{eqa_chebyshev_expansion} with the active polynomial degree scale in Eq.~\eqref{eqa_chebyshev_degree_scale}. Thus, the layer alignment of the rational dictionary elements does not by itself require the \textit{REC} model to reduce every metric relative to the \textit{Chebyshev} model on the singularly perturbed scalar BVP.

\subsection{Thermal entrance problem}\label{sec3b}

\begin{figure*}[!t]
\centerline{\includegraphics[width=1.00\textwidth]{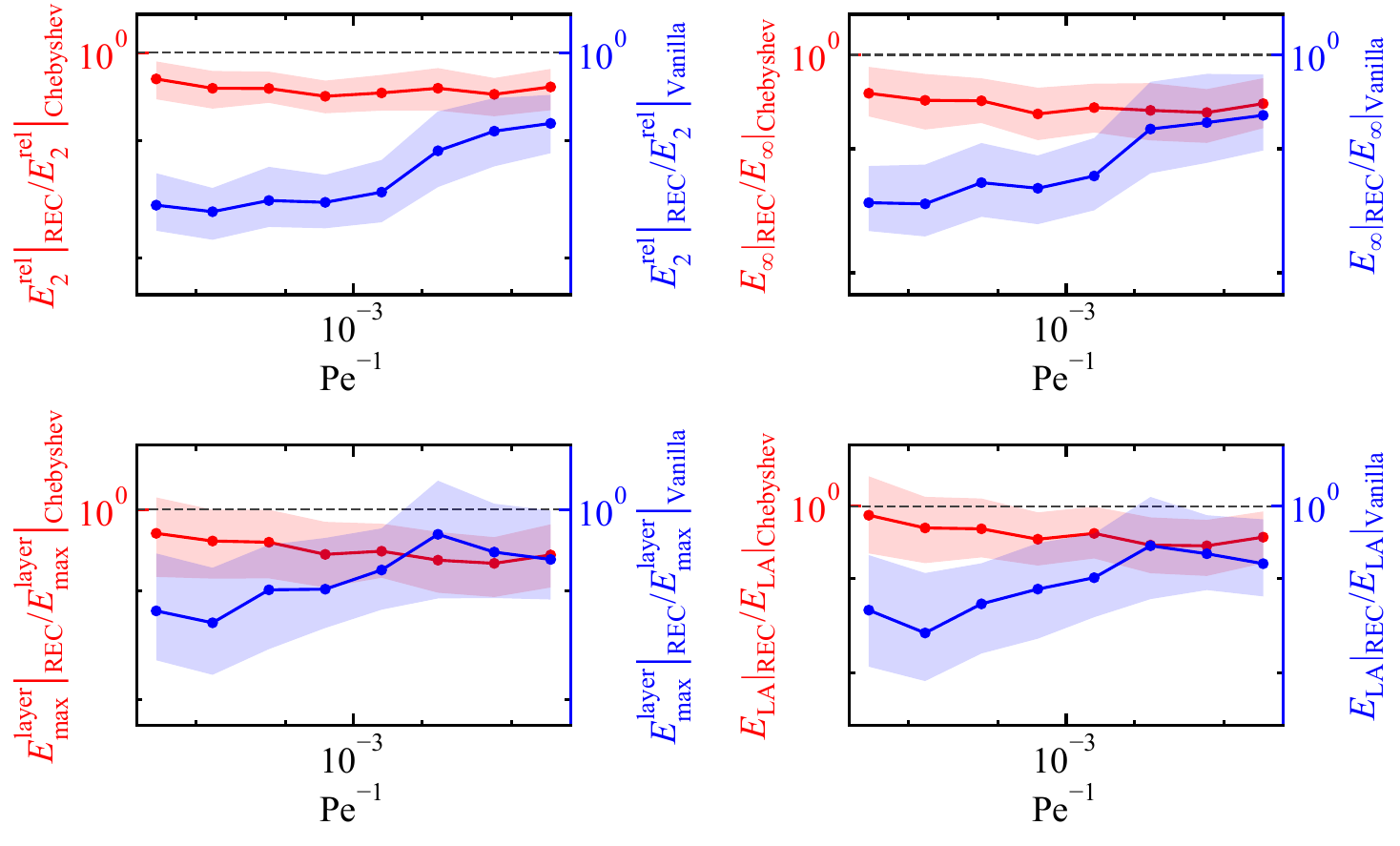}}
\caption{Ratios of $E_{2}^{\mathrm{rel}}$, $E_{\infty}$, $E_{\max}^{\mathrm{layer}}$, and $E_{\mathrm{LA}}$ over $\mathrm{Pe}^{-1}\in[10^{-4},10^{-2}]$ for the thermal entrance problem at $N_{\mathrm{out}}=16$.}
\label{fig:problem2_iqr}
\end{figure*}

Figure~\ref{fig:problem2_iqr} illustrates the ratios of the error from the \textit{REC} model to the error from the \textit{Chebyshev} model and to the error from the \textit{Vanilla} model for the thermal entrance problem, calculated in the same way as Fig.~\ref{fig:problem1_iqr}, with $\mathrm{Pe}^{-1}$ replacing $\varepsilon$ and with the independently sampled $x$ values retained within each $\mathrm{Pe}^{-1}$ bin. Relative to the \textit{Vanilla} model, the median ratios remain below unity for all eight $\mathrm{Pe}^{-1}$ bins and all four metrics, with the largest median value equal to $0.810$ for $E_{\max}^{\mathrm{layer}}$. Relative to the \textit{Chebyshev} model, the median ratios also remain below unity for all bins and metrics, with the largest median value equal to $0.918$ for $E_{\mathrm{LA}}$. These results indicate that, for the thermal entrance problem, the \textit{REC} model improves the reconstruction of the wall-normal temperature profile $\theta(y)$ across the tested $\mathrm{Pe}^{-1}$ range and sampled entrance locations $x$.

\begin{table*}[!t]
\caption{Comparison of the \textit{REC} model with the \textit{Vanilla} and \textit{Chebyshev} models for the thermal entrance problem at $N_{\mathrm{out}}=16$.}
\centering
\begingroup
\footnotesize
\setlength{\tabcolsep}{1pt}
\renewcommand{\arraystretch}{1.12}
\begin{tabular}{L{\dimexpr0.330\textwidth-2\tabcolsep\relax}C{\dimexpr0.110\textwidth-2\tabcolsep\relax}C{\dimexpr0.180\textwidth-2\tabcolsep\relax}C{\dimexpr0.190\textwidth-2\tabcolsep\relax}C{\dimexpr0.190\textwidth-2\tabcolsep\relax}}
\toprule
Metric ratio & Full test set & First three parameter bins & Seeds with lower REC error & Profiles with lower REC error \\
\midrule
$\left.E_{2}^{\mathrm{rel}}\right|_{\mathrm{REC}}/\left.E_{2}^{\mathrm{rel}}\right|_{\mathrm{Vanilla}}$ & $0.398$ & $0.316$ & $5/5$ & $98.5\%$ \\
$\left.E_{\infty}\right|_{\mathrm{REC}}/\left.E_{\infty}\right|_{\mathrm{Vanilla}}$ & $0.461$ & $0.375$ & $5/5$ & $93.7\%$ \\
$\left.E_{\max}^{\mathrm{layer}}\right|_{\mathrm{REC}}/\left.E_{\max}^{\mathrm{layer}}\right|_{\mathrm{Vanilla}}$ & $0.558$ & $0.435$ & $5/5$ & $80.6\%$ \\
$\left.E_{\mathrm{LA}}\right|_{\mathrm{REC}}/\left.E_{\mathrm{LA}}\right|_{\mathrm{Vanilla}}$ & $0.426$ & $0.318$ & $5/5$ & $85.9\%$ \\
\cmidrule(lr){1-5}
$\left.E_{2}^{\mathrm{rel}}\right|_{\mathrm{REC}}/\left.E_{2}^{\mathrm{rel}}\right|_{\mathrm{Chebyshev}}$ & $0.747$ & $0.773$ & $5/5$ & $90.6\%$ \\
$\left.E_{\infty}\right|_{\mathrm{REC}}/\left.E_{\infty}\right|_{\mathrm{Chebyshev}}$ & $0.699$ & $0.746$ & $5/5$ & $90.9\%$ \\
$\left.E_{\max}^{\mathrm{layer}}\right|_{\mathrm{REC}}/\left.E_{\max}^{\mathrm{layer}}\right|_{\mathrm{Chebyshev}}$ & $0.690$ & $0.759$ & $5/5$ & $81.2\%$ \\
$\left.E_{\mathrm{LA}}\right|_{\mathrm{REC}}/\left.E_{\mathrm{LA}}\right|_{\mathrm{Chebyshev}}$ & $0.725$ & $0.731$ & $5/5$ & $74.7\%$ \\
\bottomrule
\end{tabular}
\endgroup
\label{tab:problem2_outer16_comparison}
\end{table*}

Table~\ref{tab:problem2_outer16_comparison} summarizes error ratios for the thermal entrance problem at $N_{\mathrm{out}}=16$ for the \textit{REC} model against both the \textit{Vanilla} and \textit{Chebyshev} models, similar to Table~\ref{tab:problem1_outer16_comparison}, but with the first three parameter-bin means computed using the same parameter bins as Fig.~\ref{fig:problem2_iqr}. Relative to the \textit{Vanilla} model, all four second-column ratios are below $0.558$, and all four third-column ratios are below $0.435$. The fourth column shows that the \textit{REC} model gives lower mean error than the \textit{Vanilla} model in all five independent training runs for every metric. The fifth column remains high as well, ranging from $80.6\%$ to $98.5\%$ of the $2500$ profile comparisons. Relative to the \textit{Chebyshev} model, all four second-column ratios and all four third-column ratios remain below unity, with values between $0.690$ and $0.773$. The fourth column again shows that the \textit{REC} model gives lower mean error than the \textit{Chebyshev} model in all five independent training runs for all four metrics, and the fifth column remains between $74.7\%$ and $90.9\%$. These results indicate that, for the thermal entrance problem, the \textit{REC} model provides a more reliable representation of the wall-normal temperature profile $\theta(y)$ across sampled $\mathrm{Pe}^{-1}$ values and entrance locations $x$, where the profile must combine a smooth outer region with a wall-attached thermal layer.

\subsection{Concentration entrance problem}\label{sec3c}

\begin{figure*}[!t]
\centerline{\includegraphics[width=1.00\textwidth]{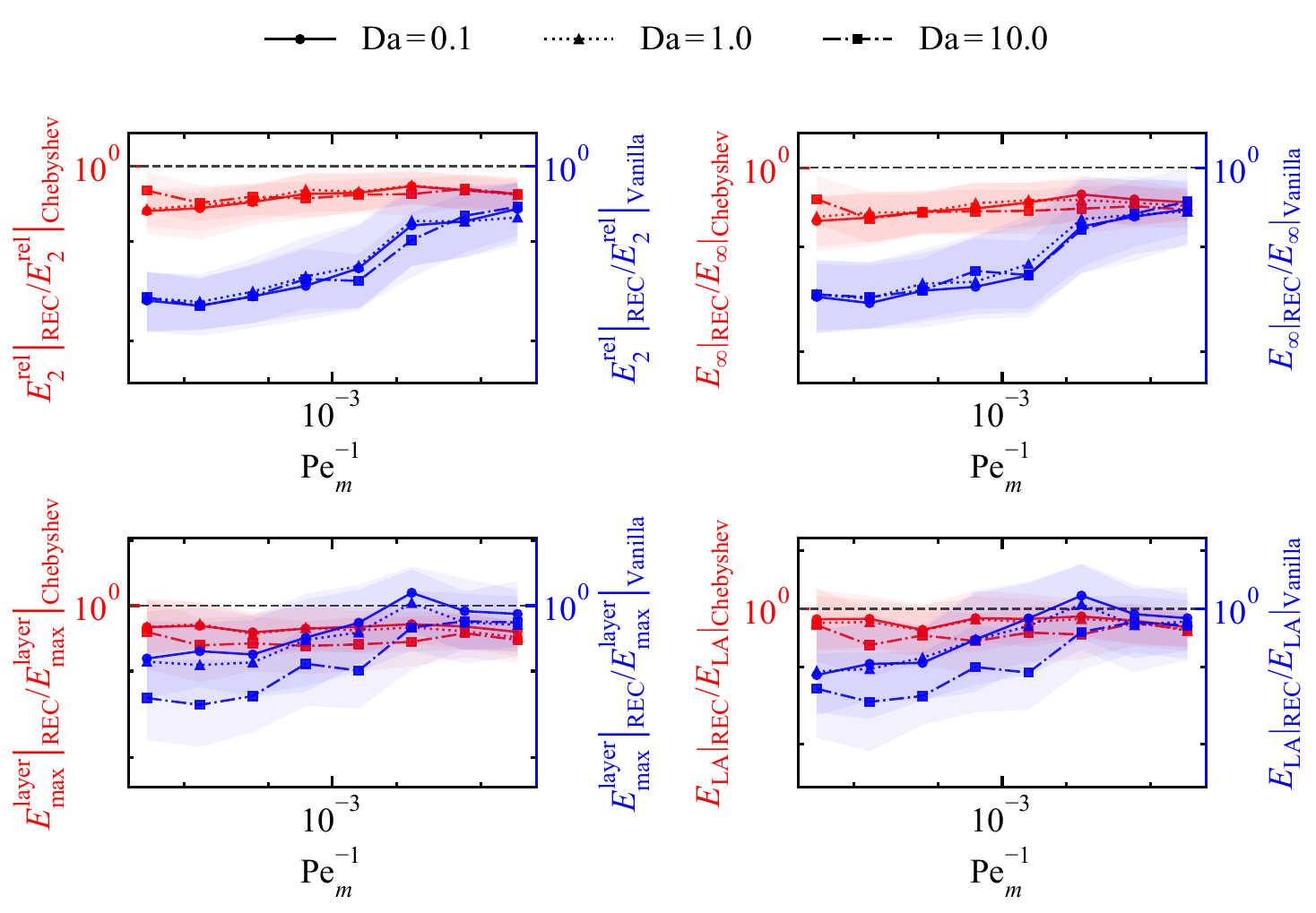}}
\caption{Ratios of $E_{2}^{\mathrm{rel}}$, $E_{\infty}$, $E_{\max}^{\mathrm{layer}}$, and $E_{\mathrm{LA}}$ over $\mathrm{Pe}_m^{-1}\in[10^{-4},10^{-2}]$ for the concentration entrance problem at $N_{\mathrm{out}}=16$ and $\mathrm{Da}=0.1$, $1.0$, and $10.0$.}
\label{fig:problem3_combined_iqr}
\end{figure*} 

Figure~\ref{fig:problem3_combined_iqr} reports the ratios of the error from the \textit{REC} model to the error from the \textit{Chebyshev} model and to the error from the \textit{Vanilla} model for the concentration entrance problem at $\mathrm{Da}=0.1$, $1.0$, and $10.0$, calculated in the same way as Fig.~\ref{fig:problem1_iqr}, with $\mathrm{Pe}_m^{-1}$ replacing $\varepsilon$ and with the independently sampled $x$ values retained within each $\mathrm{Pe}_m^{-1}$ bin. Relative to the \textit{Vanilla} model, the median ratios for $E_{2}^{\mathrm{rel}}$ and $E_{\infty}$ remain below unity for all three Damk\"ohler numbers and all eight $\mathrm{Pe}_m^{-1}$ bins, with largest median values $0.688$ and $0.745$, respectively. For $E_{\max}^{\mathrm{layer}}$ and $E_{\mathrm{LA}}$, the median ratios remain below unity in all bins except the sixth $\mathrm{Pe}_m^{-1}$ bin at $\mathrm{Da}=0.1$ and $1.0$. Relative to the \textit{Chebyshev} model, all median ratios remain below unity for all three Damk\"ohler numbers, all eight bins, and all four metrics, with the largest median value equal to $0.914$ for $E_{\mathrm{LA}}$. These results indicate that, for the concentration entrance problem, the \textit{REC} model improves the reconstruction of the wall-normal concentration profile $c(y)$ across sampled entrance locations $x$ and across wall-absorption regimes ranging from weak uptake at $\mathrm{Da}=0.1$ to stronger wall-adjacent concentration gradients at $\mathrm{Da}=10.0$.

\begin{table*}[!tp]
\caption{Comparison of the \textit{REC} model with the \textit{Vanilla} and \textit{Chebyshev} models for the concentration entrance problem at $N_{\mathrm{out}}=16$.}
\centering
\begingroup
\footnotesize
\setlength{\tabcolsep}{1pt}
\renewcommand{\arraystretch}{1.12}
\begin{tabular}{L{\dimexpr0.330\textwidth-2\tabcolsep\relax}C{\dimexpr0.110\textwidth-2\tabcolsep\relax}C{\dimexpr0.180\textwidth-2\tabcolsep\relax}C{\dimexpr0.190\textwidth-2\tabcolsep\relax}C{\dimexpr0.190\textwidth-2\tabcolsep\relax}}
\toprule
Metric ratio & Full test set & First three parameter bins & Seeds with lower REC error & Profiles with lower REC error \\
\midrule
\multicolumn{5}{@{}l}{Problem~3 ($\mathrm{Da}=0.1$)} \\
$\left.E_{2}^{\mathrm{rel}}\right|_{\mathrm{REC}}/\left.E_{2}^{\mathrm{rel}}\right|_{\mathrm{Vanilla}}$ & $0.453$ & $0.291$ & $5/5$ & $94.9\%$ \\
$\left.E_{\infty}\right|_{\mathrm{REC}}/\left.E_{\infty}\right|_{\mathrm{Vanilla}}$ & $0.476$ & $0.329$ & $5/5$ & $90.9\%$ \\
$\left.E_{\max}^{\mathrm{layer}}\right|_{\mathrm{REC}}/\left.E_{\max}^{\mathrm{layer}}\right|_{\mathrm{Vanilla}}$ & $0.799$ & $0.585$ & $5/5$ & $64.2\%$ \\
$\left.E_{\mathrm{LA}}\right|_{\mathrm{REC}}/\left.E_{\mathrm{LA}}\right|_{\mathrm{Vanilla}}$ & $0.827$ & $0.578$ & $4/5$ & $63.8\%$ \\
\cmidrule(lr){1-5}
$\left.E_{2}^{\mathrm{rel}}\right|_{\mathrm{REC}}/\left.E_{2}^{\mathrm{rel}}\right|_{\mathrm{Chebyshev}}$ & $0.761$ & $0.683$ & $5/5$ & $87.7\%$ \\
$\left.E_{\infty}\right|_{\mathrm{REC}}/\left.E_{\infty}\right|_{\mathrm{Chebyshev}}$ & $0.720$ & $0.661$ & $5/5$ & $90.4\%$ \\
$\left.E_{\max}^{\mathrm{layer}}\right|_{\mathrm{REC}}/\left.E_{\max}^{\mathrm{layer}}\right|_{\mathrm{Chebyshev}}$ & $0.788$ & $0.784$ & $5/5$ & $77.7\%$ \\
$\left.E_{\mathrm{LA}}\right|_{\mathrm{REC}}/\left.E_{\mathrm{LA}}\right|_{\mathrm{Chebyshev}}$ & $0.895$ & $0.943$ & $5/5$ & $65.5\%$ \\
\midrule
\multicolumn{5}{@{}l}{Problem~3 ($\mathrm{Da}=1.0$)} \\
$\left.E_{2}^{\mathrm{rel}}\right|_{\mathrm{REC}}/\left.E_{2}^{\mathrm{rel}}\right|_{\mathrm{Vanilla}}$ & $0.463$ & $0.304$ & $5/5$ & $93.5\%$ \\
$\left.E_{\infty}\right|_{\mathrm{REC}}/\left.E_{\infty}\right|_{\mathrm{Vanilla}}$ & $0.484$ & $0.339$ & $5/5$ & $90.8\%$ \\
$\left.E_{\max}^{\mathrm{layer}}\right|_{\mathrm{REC}}/\left.E_{\max}^{\mathrm{layer}}\right|_{\mathrm{Vanilla}}$ & $0.757$ & $0.564$ & $5/5$ & $69.3\%$ \\
$\left.E_{\mathrm{LA}}\right|_{\mathrm{REC}}/\left.E_{\mathrm{LA}}\right|_{\mathrm{Vanilla}}$ & $0.829$ & $0.620$ & $4/5$ & $68\%$ \\
\cmidrule(lr){1-5}
$\left.E_{2}^{\mathrm{rel}}\right|_{\mathrm{REC}}/\left.E_{2}^{\mathrm{rel}}\right|_{\mathrm{Chebyshev}}$ & $0.770$ & $0.702$ & $5/5$ & $87.8\%$ \\
$\left.E_{\infty}\right|_{\mathrm{REC}}/\left.E_{\infty}\right|_{\mathrm{Chebyshev}}$ & $0.720$ & $0.670$ & $5/5$ & $91.6\%$ \\
$\left.E_{\max}^{\mathrm{layer}}\right|_{\mathrm{REC}}/\left.E_{\max}^{\mathrm{layer}}\right|_{\mathrm{Chebyshev}}$ & $0.774$ & $0.782$ & $5/5$ & $79.5\%$ \\
$\left.E_{\mathrm{LA}}\right|_{\mathrm{REC}}/\left.E_{\mathrm{LA}}\right|_{\mathrm{Chebyshev}}$ & $0.891$ & $0.954$ & $5/5$ & $68.8\%$ \\
\midrule
\multicolumn{5}{@{}l}{Problem~3 ($\mathrm{Da}=10.0$)} \\
$\left.E_{2}^{\mathrm{rel}}\right|_{\mathrm{REC}}/\left.E_{2}^{\mathrm{rel}}\right|_{\mathrm{Vanilla}}$ & $0.433$ & $0.307$ & $5/5$ & $96.1\%$ \\
$\left.E_{\infty}\right|_{\mathrm{REC}}/\left.E_{\infty}\right|_{\mathrm{Vanilla}}$ & $0.467$ & $0.350$ & $5/5$ & $92.7\%$ \\
$\left.E_{\max}^{\mathrm{layer}}\right|_{\mathrm{REC}}/\left.E_{\max}^{\mathrm{layer}}\right|_{\mathrm{Vanilla}}$ & $0.584$ & $0.388$ & $5/5$ & $79.3\%$ \\
$\left.E_{\mathrm{LA}}\right|_{\mathrm{REC}}/\left.E_{\mathrm{LA}}\right|_{\mathrm{Vanilla}}$ & $0.652$ & $0.518$ & $5/5$ & $79.2\%$ \\
\cmidrule(lr){1-5}
$\left.E_{2}^{\mathrm{rel}}\right|_{\mathrm{REC}}/\left.E_{2}^{\mathrm{rel}}\right|_{\mathrm{Chebyshev}}$ & $0.760$ & $0.754$ & $5/5$ & $87.9\%$ \\
$\left.E_{\infty}\right|_{\mathrm{REC}}/\left.E_{\infty}\right|_{\mathrm{Chebyshev}}$ & $0.698$ & $0.705$ & $5/5$ & $90.3\%$ \\
$\left.E_{\max}^{\mathrm{layer}}\right|_{\mathrm{REC}}/\left.E_{\max}^{\mathrm{layer}}\right|_{\mathrm{Chebyshev}}$ & $0.678$ & $0.681$ & $5/5$ & $82.9\%$ \\
$\left.E_{\mathrm{LA}}\right|_{\mathrm{REC}}/\left.E_{\mathrm{LA}}\right|_{\mathrm{Chebyshev}}$ & $0.710$ & $0.656$ & $5/5$ & $72.8\%$ \\
\bottomrule
\end{tabular}
\endgroup
\label{tab:problem3_outer16_comparison}
\end{table*}

Table~\ref{tab:problem3_outer16_comparison} lists error ratios for the concentration entrance problem at $N_{\mathrm{out}}=16$ and $\mathrm{Da}=0.1$, $1.0$, and $10.0$ for the \textit{REC} model against both the \textit{Vanilla} and \textit{Chebyshev} models, similar to Table~\ref{tab:problem1_outer16_comparison}, but with the first three parameter-bin means computed using the same parameter bins as Fig.~\ref{fig:problem3_combined_iqr}. Relative to the \textit{Vanilla} model, the second-column ratios are below unity for all three Damk\"ohler numbers and all four metrics, and the third-column ratios are smaller than the corresponding second-column ratios in every row. The fourth column shows that the \textit{REC} model gives lower mean error than the \textit{Vanilla} model in all five independent training runs for every metric except $E_{\mathrm{LA}}$ at $\mathrm{Da}=0.1$ and $1.0$, where this occurs in four of the five independent training runs, while the fifth column remains above $63.8\%$ across all rows. Relative to the \textit{Chebyshev} model, all second-column and third-column ratios remain below unity for $\mathrm{Da}=0.1$, $1.0$, and $10.0$. The fourth column shows that the \textit{REC} model gives lower mean error than the \textit{Chebyshev} model in all five independent training runs for each of the four metrics at each tested Damk\"ohler number, and the fifth column ranges from $65.5\%$ to $91.6\%$. These results indicate that, for the concentration entrance problem, the \textit{REC} model remains effective for reconstructing the wall-normal concentration profile $c(y)$ as the absorbing-wall condition changes the near-wall layer from weakly developed to sharply localized.

\subsection{Representative profile behavior and local smoothness}\label{sec3d}

\begin{sidewaysfigure*}[p]
\centering
\includegraphics[width=1.00\textheight]{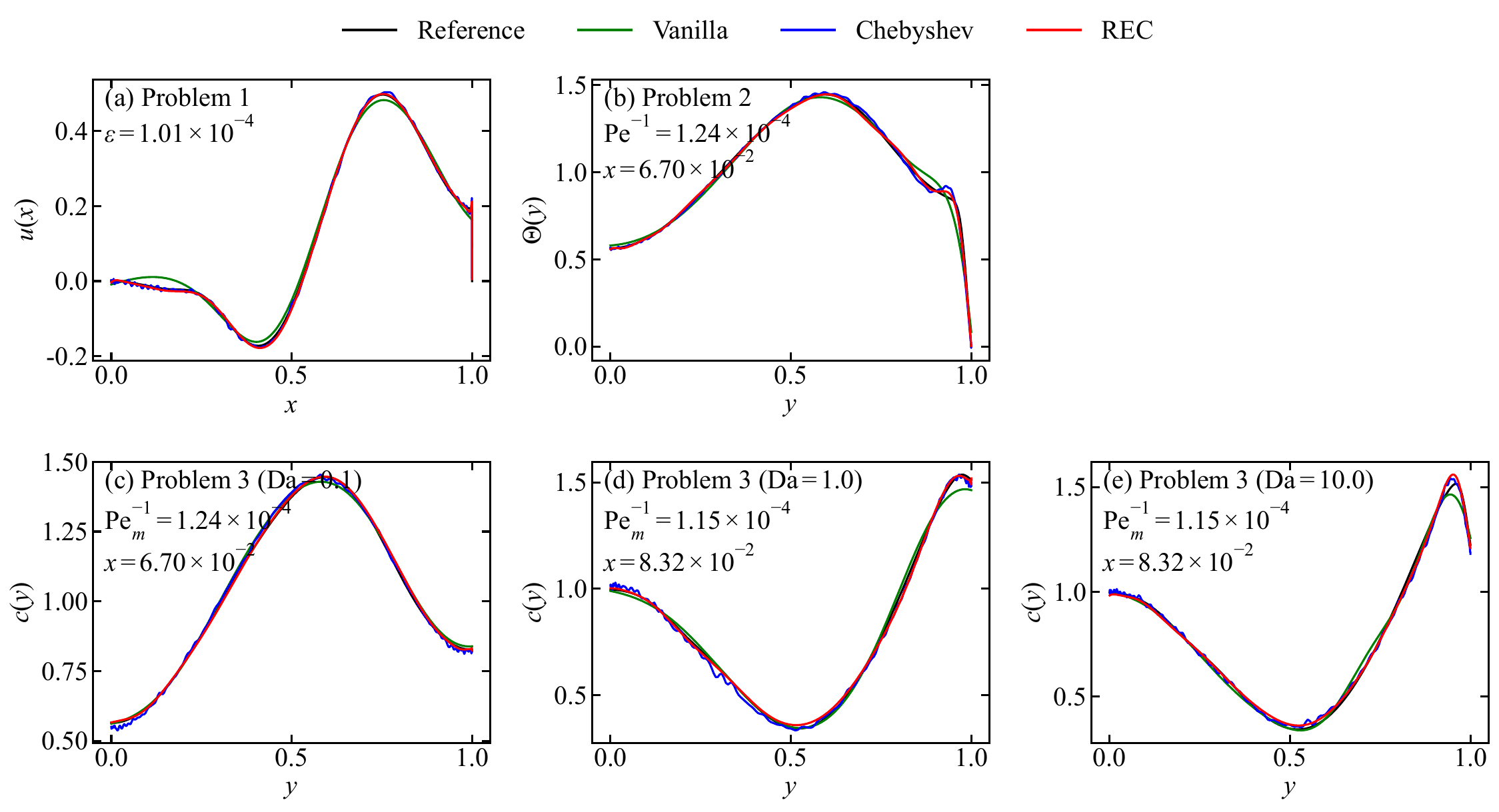}
\caption{Full representative solution profiles for the three problems at $N_{\mathrm{out}}=16$. (a) Scalar profile $u(x)$ on $x\in(0,1)$ for $\varepsilon=1.01\times10^{-4}$. (b) Temperature profile $\theta(y)$ on $y\in(0,1)$ for $\mathrm{Pe}^{-1}=1.24\times10^{-4}$ and $x=6.70\times10^{-2}$. (c)--(e) Concentration profiles $c(y)$ on $y\in(0,1)$ for $\mathrm{Da}=0.1$, $1.0$, and $10.0$, where the $\mathrm{Da}=0.1$ case uses $\mathrm{Pe}_m^{-1}=1.24\times10^{-4}$ and $x=6.70\times10^{-2}$, while the $\mathrm{Da}=1.0$ and $10.0$ cases use $\mathrm{Pe}_m^{-1}=1.15\times10^{-4}$ and $x=8.32\times10^{-2}$.}
\label{fig:overview_full}
\end{sidewaysfigure*}

Figure~\ref{fig:overview_full} shows representative cases selected from the held-out test profiles to visualize the profile behavior underlying the error statistics reported in Sections~\ref{sec3a}--\ref{sec3c}. The figure compares the numerical reference with solution profiles predicted by the \textit{Vanilla}, \textit{Chebyshev}, and \textit{REC} models at $N_{\mathrm{out}}=16$ across the three benchmark problems introduced in Section~\ref{sec2a}. For Problems~1 and~2, Figs.~\ref{fig:overview_full}(a) and \ref{fig:overview_full}(b) show the scalar solution $u(x)$ over $x\in(0,1)$ and the wall-normal temperature profile $\theta(y)$ over $y\in(0,1)$, respectively, with $\varepsilon=1.01\times10^{-4}$ in Fig.~\ref{fig:overview_full}(a) and $\mathrm{Pe}^{-1}=1.24\times10^{-4}$ and $x=6.70\times10^{-2}$ in Fig.~\ref{fig:overview_full}(b). In Fig.~\ref{fig:overview_full}(a), using the \textit{REC} model decreases $E_{2}^{\mathrm{rel}}$ from $7.90\times10^{-2}$ with the \textit{Vanilla} model and $2.49\times10^{-2}$ with the \textit{Chebyshev} model to $1.87\times10^{-2}$, and decreases $E_{\infty}$ from $1.63\times10^{-1}$ and $3.03\times10^{-2}$ to $2.15\times10^{-2}$. In Fig.~\ref{fig:overview_full}(b), $E_{2}^{\mathrm{rel}}$ decreases from $4.45\times10^{-2}$ with the \textit{Vanilla} model and $2.41\times10^{-2}$ with the \textit{Chebyshev} model to $1.63\times10^{-2}$, and $E_{\infty}$ decreases from $1.59\times10^{-1}$ and $8.01\times10^{-2}$ to $5.91\times10^{-2}$. These reductions show that the \textit{REC} model is much closer to the numerical reference than the \textit{Vanilla} model and still improves over the \textit{Chebyshev} model in the representative scalar and thermal entrance profiles.

For Problem~3, Figs.~\ref{fig:overview_full}(c), \ref{fig:overview_full}(d), and \ref{fig:overview_full}(e) show the wall-normal concentration profiles $c(y)$ over $y\in(0,1)$ as the wall-absorption strength $\mathrm{Da}$ increases from $0.1$ to $10.0$, with $\mathrm{Pe}_m^{-1}=1.24\times10^{-4}$ and $x=6.70\times10^{-2}$ for $\mathrm{Da}=0.1$, and $\mathrm{Pe}_m^{-1}=1.15\times10^{-4}$ and $x=8.32\times10^{-2}$ for $\mathrm{Da}=1.0$ and $10.0$. The wall-endpoint drop of the numerical reference, measured by $c(y=0.99)-c(y=1)$, increases from $9.44\times10^{-4}$ at $\mathrm{Da}=0.1$ to $1.43\times10^{-2}$ at $\mathrm{Da}=1.0$ and $1.19\times10^{-1}$ at $\mathrm{Da}=10.0$, showing the progressive formation of a sharper wall-adjacent concentration layer as $\mathrm{Da}$ increases. Relative to the \textit{Vanilla} model, the \textit{REC} model decreases $E_{2}^{\mathrm{rel}}$ from $1.13\times10^{-2}$, $2.74\times10^{-2}$, and $2.48\times10^{-2}$ to $5.63\times10^{-3}$, $1.22\times10^{-2}$, and $1.82\times10^{-2}$, respectively, and decreases $E_{\infty}$ from $2.30\times10^{-2}$, $7.00\times10^{-2}$, and $7.64\times10^{-2}$ to $1.56\times10^{-2}$, $3.10\times10^{-2}$, and $5.69\times10^{-2}$. Relative to the \textit{Chebyshev} model, $E_{2}^{\mathrm{rel}}$ decreases from $1.17\times10^{-2}$, $1.88\times10^{-2}$, and $2.16\times10^{-2}$ to the same \textit{REC} values. For $\mathrm{Da}=0.1$ and $1.0$, $E_{\infty}$ decreases from $2.83\times10^{-2}$ to $1.56\times10^{-2}$ and from $5.88\times10^{-2}$ to $3.10\times10^{-2}$, while the $\mathrm{Da}=10.0$ value remains $5.69\times10^{-2}$ for both models. Even at $\mathrm{Da}=0.1$, where the wall-endpoint drop is weak, the \textit{Chebyshev} model still introduces visible near-wall oscillation, whereas the \textit{REC} model follows the nearly flat reference profile more closely. These results indicate that the \textit{REC} model predicts the numerical reference substantially more accurately than the \textit{Vanilla} model and also improves the prediction relative to the \textit{Chebyshev} model by reducing near-wall oscillatory deviations, rather than only by resolving an extremely thin localized layer.

\begin{sidewaysfigure*}[p]
\centering
\includegraphics[width=1.00\textheight]{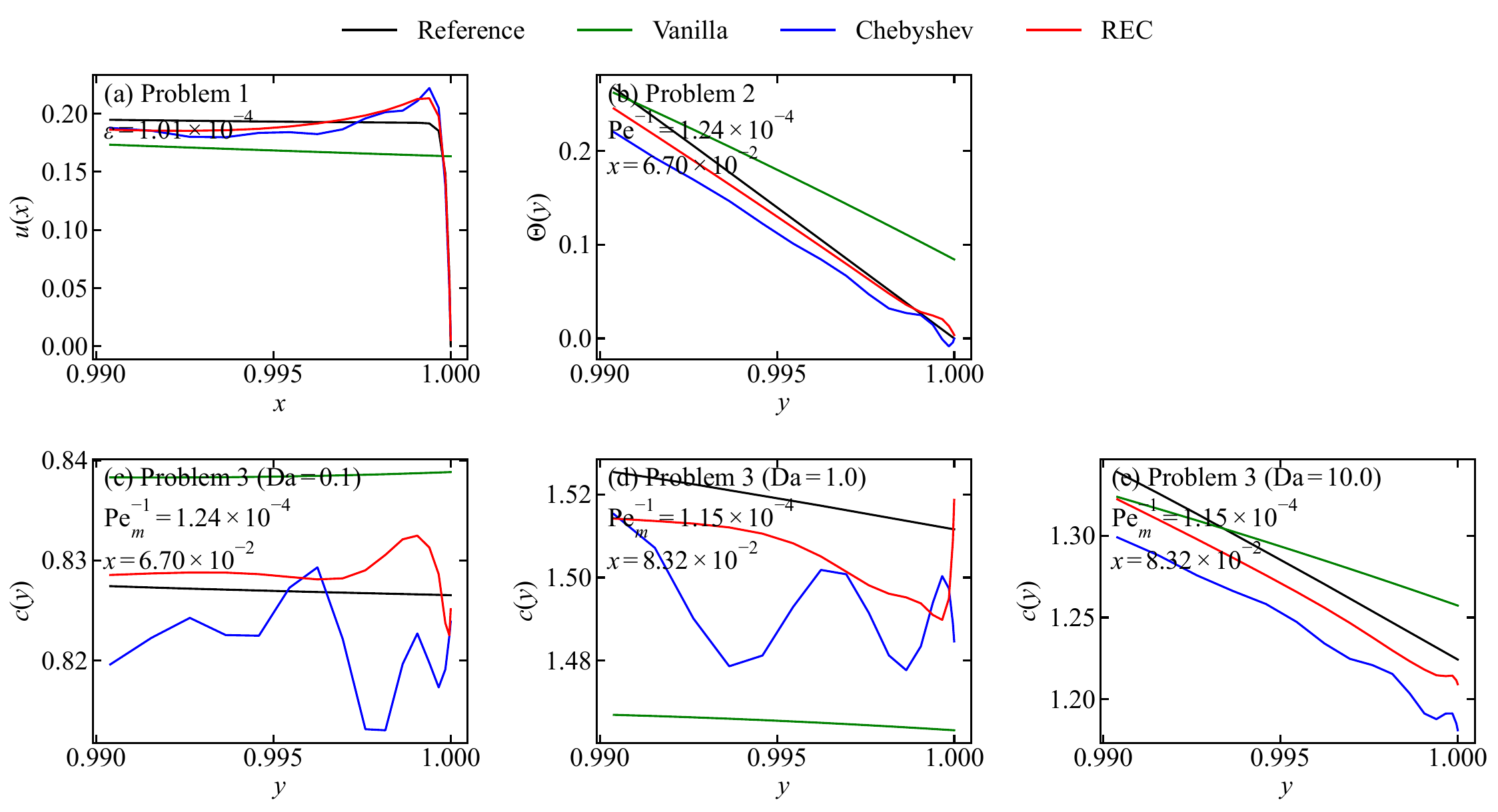}
\caption{Near-boundary and near-wall solution profiles for the three problems at $N_{\mathrm{out}}=16$. (a) Scalar profile $u(x)$ near the right endpoint for $\varepsilon=1.01\times10^{-4}$. (b) Temperature profile $\theta(y)$ near the wall for $\mathrm{Pe}^{-1}=1.24\times10^{-4}$ and $x=6.70\times10^{-2}$. (c)--(e) Concentration profiles $c(y)$ near the absorbing wall for $\mathrm{Da}=0.1$, $1.0$, and $10.0$, where the $\mathrm{Da}=0.1$ case uses $\mathrm{Pe}_m^{-1}=1.24\times10^{-4}$ and $x=6.70\times10^{-2}$, while the $\mathrm{Da}=1.0$ and $10.0$ cases use $\mathrm{Pe}_m^{-1}=1.15\times10^{-4}$ and $x=8.32\times10^{-2}$. Each comparison includes the numerical reference, Vanilla, Chebyshev, and REC.}
\label{fig:overview_zoom}
\end{sidewaysfigure*}

Figure~\ref{fig:overview_zoom} enlarges the right-endpoint layer in $u(x)$, the near-wall part of the temperature profile $\theta(y)$, and the near-wall part of the concentration profiles $c(y)$ at $\mathrm{Da}=0.1$, $1.0$, and $10.0$ for the five representative cases shown in Fig.~\ref{fig:overview_full}. In these enlarged regions, the predictions from the \textit{Chebyshev} model often retain local oscillatory deviations from the numerical reference, whereas the predictions from the \textit{REC} model reduce these deviations while preserving the nearby smooth profile shape. To quantify this local oscillatory behavior, the second-difference roughness $W_2$ is evaluated for the same representative profiles. Profiles with stronger oscillatory variation have larger second differences, while $W_2$ values closer to the numerical-reference value indicate closer agreement in local smoothness. For a discrete profile vector $\mathbf v=(v_1,\ldots,v_m)$ evaluated at the $m=257$ Chebyshev--Lobatto output locations defined in Section~\ref{sec2d}, with $v_j=u(x_j)$ for Problem~1, $v_j=\theta(y_j)$ for Problem~2, and $v_j=c(y_j)$ for Problem~3, where $j$ indexes the output locations,
\begin{equation}
W_2(\mathbf v)=\sum_{j=2}^{m-1}\left|v_{j+1}-2v_j+v_{j-1}\right|,
\end{equation}
where $|\cdot|$ denotes the scalar absolute value.

\begin{table*}[t]
\caption{$W_2$ values for the numerical reference and for the predictions from the \textit{Chebyshev} and \textit{REC} models in the five representative cases shown in Fig.~\ref{fig:overview_zoom}, computed over $x\in(0,1)$ for Problem~1 and over $y\in(0,1)$ for Problems~2 and~3.}
\centering
\begingroup
\footnotesize
\setlength{\tabcolsep}{2pt}
\renewcommand{\arraystretch}{1.05}
\begin{tabular}{L{\dimexpr0.170\textwidth-2\tabcolsep\relax}L{\dimexpr0.265\textwidth-2\tabcolsep\relax}C{\dimexpr0.190\textwidth-2\tabcolsep\relax}C{\dimexpr0.190\textwidth-2\tabcolsep\relax}C{\dimexpr0.185\textwidth-2\tabcolsep\relax}}
\toprule
Problem & Parameter & $\left.W_2\right|_{\mathrm{Reference}}$ & $\left.W_2\right|_{\mathrm{Chebyshev}}$ & $\left.W_2\right|_{\mathrm{REC}}$ \\
\midrule
Problem 1 & $\varepsilon = 1.01 \times 10^{-4}$ & $1.87 \times 10^{-1}$ & $9.72 \times 10^{-1}$ & $2.07 \times 10^{-1}$ \\
\cmidrule(lr){1-5}
Problem 2 & \shortstack[l]{$\mathrm{Pe}^{-1} = 1.24 \times 10^{-4}$\\$x=6.70 \times 10^{-2}$} & $1.29 \times 10^{-1}$ & $8.23 \times 10^{-1}$ & $1.35 \times 10^{-1}$ \\
\cmidrule(lr){1-5}
\shortstack[l]{Problem 3\\($\mathrm{Da}=0.1$)} & \shortstack[l]{$\mathrm{Pe}_m^{-1} = 1.24 \times 10^{-4}$\\$x=6.70 \times 10^{-2}$} & $5.15 \times 10^{-2}$ & $8.71 \times 10^{-1}$ & $6.78 \times 10^{-2}$ \\
\shortstack[l]{Problem 3\\($\mathrm{Da}=1.0$)} & \shortstack[l]{$\mathrm{Pe}_m^{-1} = 1.15 \times 10^{-4}$\\$x=8.32 \times 10^{-2}$} & $6.39 \times 10^{-2}$ & $1.43 \times 10^{0}$ & $9.26 \times 10^{-2}$ \\
\shortstack[l]{Problem 3\\($\mathrm{Da}=10.0$)} & \shortstack[l]{$\mathrm{Pe}_m^{-1} = 1.15 \times 10^{-4}$\\$x=8.32 \times 10^{-2}$} & $9.22 \times 10^{-2}$ & $1.63 \times 10^{0}$ & $9.63 \times 10^{-2}$ \\
\bottomrule
\end{tabular}
\endgroup
\label{tab:roughness}
\end{table*}

Table~\ref{tab:roughness} shows that the $W_2$ values from the \textit{REC} model remain closer to the numerical-reference values than the $W_2$ values from the \textit{Chebyshev} model in all five representative cases. For the thermal entrance problem, $W_2$ decreases by $83.6\%$ from the \textit{Chebyshev} model to the \textit{REC} model. For the concentration entrance problem, the corresponding decreases are $92.2\%$, $93.5\%$, and $94.1\%$ at $\mathrm{Da}=0.1$, $1.0$, and $10.0$, respectively, giving larger decreases than in the thermal entrance problem. More importantly, for the singularly perturbed scalar BVP, although the \textit{REC} model does not substantially improve the scalar-profile prediction accuracy relative to the \textit{Chebyshev} model, it decreases $W_2$ by $78.7\%$. In addition, for all held-out test profiles over the five independent training runs, the median $W_2$ decreases from the \textit{Chebyshev} model to the \textit{REC} model by $64.0\%$ for the singularly perturbed scalar BVP and by $81.4\%$ for the thermal entrance problem. For the concentration entrance problem, the corresponding median decreases are $87.2\%$, $87.6\%$, and $85.9\%$ at $\mathrm{Da}=0.1$, $1.0$, and $10.0$, respectively. These results indicate that the \textit{REC} model suppresses the artificial local reversals introduced by the \textit{Chebyshev} model near the boundary or wall, thereby giving a near-boundary or near-wall profile shape closer to the numerical-reference behavior across both the five representative cases and all held-out test profiles.

\subsection{Dependence on outer Chebyshev subdictionary size}\label{sec3e}

Sections~\ref{sec3a}--\ref{sec3d} show that, first, compared with the \textit{Vanilla} model, the \textit{REC} model gives lower errors in predicting $u(x)$, $\theta(y)$, and $c(y)$, even though the \textit{Vanilla} model uses a learned, parameter-conditioned trunk MLP. This result indicates that the learned coordinate trunk does not by itself provide the same layer-aligned output functions as the prescribed REC dictionary for the tested bounded-domain solution profiles. Compared with the \textit{Chebyshev} model, the \textit{REC} model changes the prescribed trunk dictionary while keeping the branch network, trunk dimension, data splits, loss function, optimizer, training procedure, and evaluation protocol fixed, as described in Section~\ref{sec2}. Therefore, the lower errors in the thermal and concentration entrance problems, together with the smaller $W_2$ values in the representative profiles, support replacing part of the global Chebyshev dictionary by inner-scale rational dictionary elements in the output-coordinate representation, rather than adding trainable trunk expressivity, changing the optimization procedure, or imposing a problem-specific decomposition, rescaling, or sampling rule. However, the results in Sections~\ref{sec3a}--\ref{sec3d} are obtained with the specific split $N_{\mathrm{out}}=16$ and do not yet show whether the lower errors in predicting $\theta(y)$ and $c(y)$ can be attributed to using a larger number of inner rational dictionary elements within the $p=129$ trunk dictionary, or whether similar errors would be obtained with a larger outer Chebyshev subdictionary. They also do not show whether the lower errors relative to the \textit{Vanilla} model remain when $N_{\mathrm{out}}$ is increased to $33$, $65$, and $97$.

\begin{figure*}[!t]
\centerline{\includegraphics[width=1.00\textwidth]{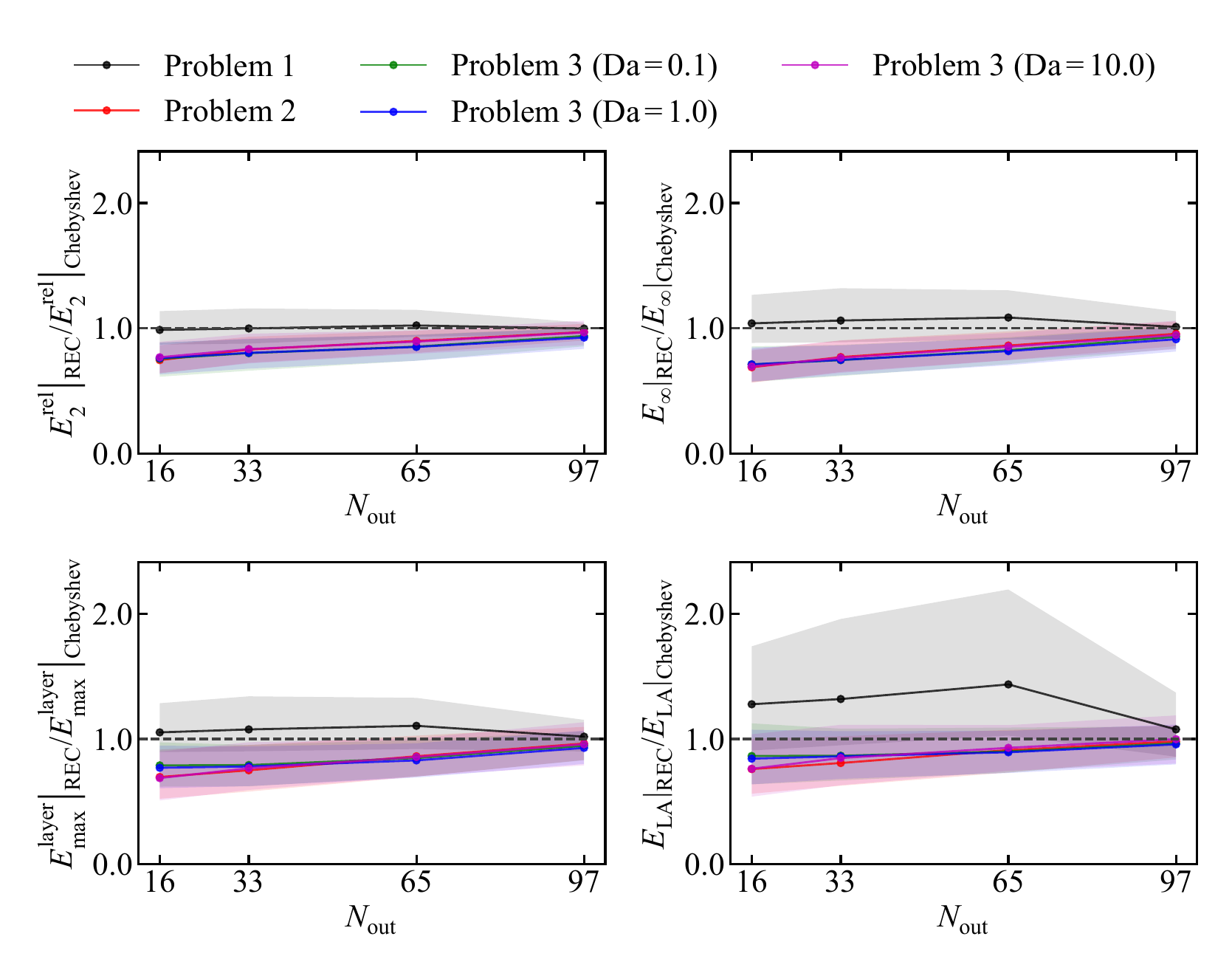}}
\caption{Ratios of $E_{2}^{\mathrm{rel}}$, $E_{\infty}$, $E_{\max}^{\mathrm{layer}}$, and $E_{\mathrm{LA}}$ from the \textit{REC} model to the corresponding errors from the \textit{Chebyshev} model for $N_{\mathrm{out}}=16$, $33$, $65$, and $97$.}
\label{fig:outer_sweep_cheb}
\end{figure*}

\begin{figure*}[!t]
\centerline{\includegraphics[width=1.00\textwidth]{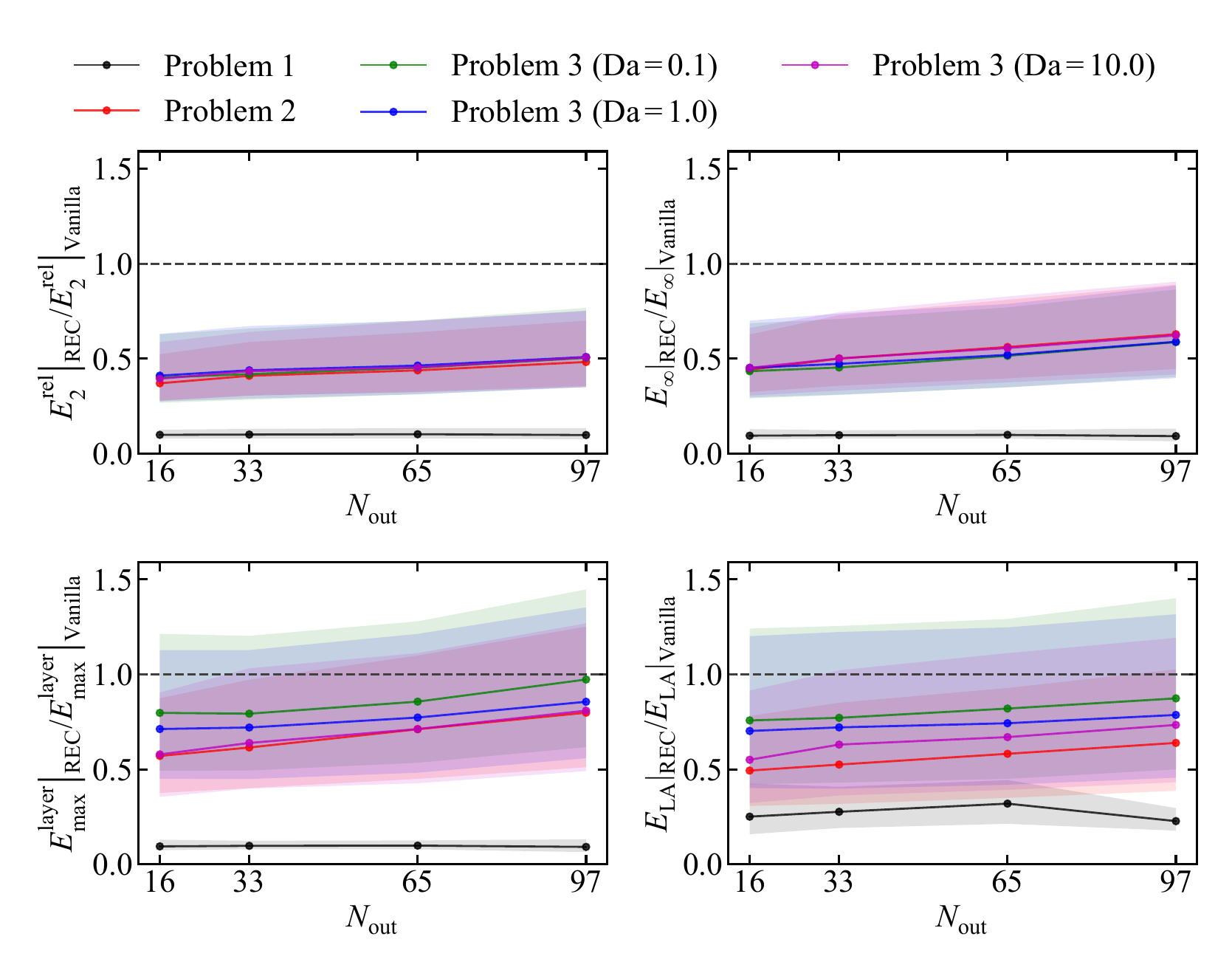}}
\caption{Ratios of $E_{2}^{\mathrm{rel}}$, $E_{\infty}$, $E_{\max}^{\mathrm{layer}}$, and $E_{\mathrm{LA}}$ from the \textit{REC} model to the corresponding errors from the \textit{Vanilla} model for $N_{\mathrm{out}}=16$, $33$, $65$, and $97$.}
\label{fig:outer_sweep_vanilla}
\end{figure*}

Figures~\ref{fig:outer_sweep_cheb} and~\ref{fig:outer_sweep_vanilla} accordingly compare the error from the \textit{REC} model against the errors from the \textit{Chebyshev} and \textit{Vanilla} models as $N_{\mathrm{out}}$ is varied over $16$, $33$, $65$, and $97$. For each metric and each problem, the solid line gives the median ratio over the five independent training runs and test profiles, and the shaded band gives the IQR. In Fig.~\ref{fig:outer_sweep_cheb}, for the singularly perturbed scalar BVP, the ratios remain close to unity and can exceed unity for $E_{\max}^{\mathrm{layer}}$ and $E_{\mathrm{LA}}$, consistent with Table~\ref{tab:problem1_outer16_comparison}, where the \textit{Chebyshev} model gives lower errors than the \textit{REC} model on these two metrics. For the thermal and concentration entrance problems in Fig.~\ref{fig:outer_sweep_cheb}, the smallest ratios occur at $N_{\mathrm{out}}=16$, and increasing $N_{\mathrm{out}}$ moves the ratios toward unity. In Fig.~\ref{fig:outer_sweep_vanilla}, the ratios remain below unity for the thermal and concentration entrance problems and remain below unity for most singularly perturbed scalar BVP cases. Thus, the weaker improvement over the \textit{Chebyshev} model at larger $N_{\mathrm{out}}$ supports the interpretation that retaining more inner rational dictionary elements is more favorable for predicting $\theta(y)$ and $c(y)$ than increasing the number of outer Chebyshev dictionary elements. The ratios in Fig.~\ref{fig:outer_sweep_vanilla} also show that the lower errors relative to the \textit{Vanilla} model are retained when $N_{\mathrm{out}}$ is increased.

\subsection{Practical implications}\label{sec3f}

\begin{figure*}[!t]
\centerline{\includegraphics[width=1.0\textwidth]{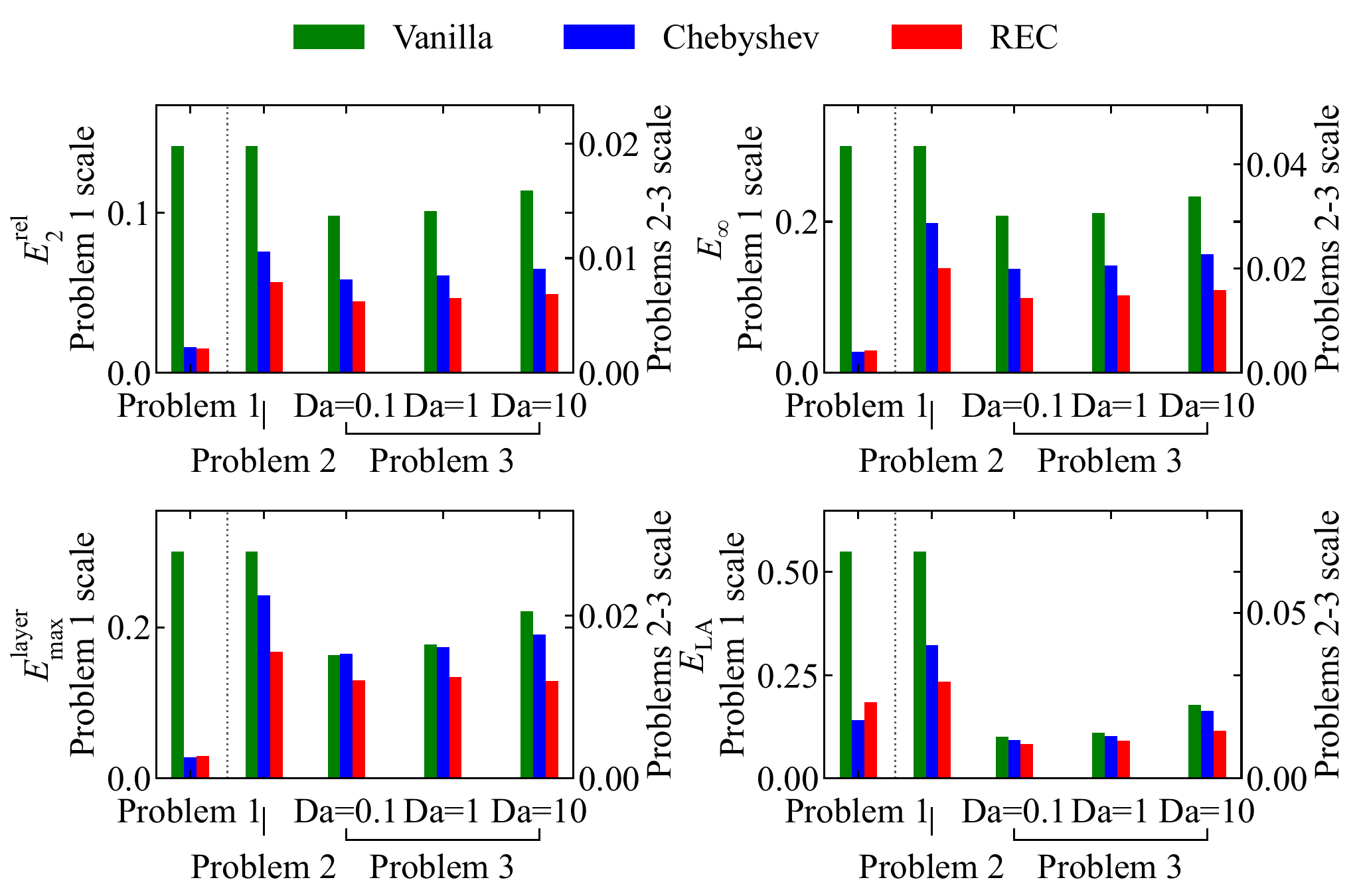}}
\caption{Averaged values of $E_{2}^{\mathrm{rel}}$, $E_{\infty}$, $E_{\max}^{\mathrm{layer}}$, and $E_{\mathrm{LA}}$ for the \textit{Vanilla}, \textit{Chebyshev}, and \textit{REC} models at $N_{\mathrm{out}}=16$.}
\label{fig:cross_metrics}
\end{figure*}

Figure~\ref{fig:cross_metrics} compares the averaged values of $E_{2}^{\mathrm{rel}}$, $E_{\infty}$, $E_{\max}^{\mathrm{layer}}$, and $E_{\mathrm{LA}}$ for the \textit{Vanilla}, \textit{Chebyshev}, and \textit{REC} models at $N_{\mathrm{out}}=16$. For each model and each problem, the plotted value is the corresponding error metric averaged over the five independent training runs and the test profiles. These averaged error values summarize the observations in Sections~\ref{sec3a}--\ref{sec3d}. For the singularly perturbed scalar BVP, the \textit{REC} model predicts the scalar profile $u(x)$ more accurately than the \textit{Vanilla} model and remains comparable to the \textit{Chebyshev} model. For the thermal and concentration entrance problems, the \textit{REC} model gives more accurate predictions of the temperature profile $\theta(y)$ and the concentration profile $c(y)$ than both comparison models. The $N_{\mathrm{out}}$ sweep in Section~\ref{sec3e} further shows that, in these two entrance problems, the lower errors in predicting $\theta(y)$ and $c(y)$ are strengthened by using more inner rational dictionary elements within the REC dictionary. Therefore, the practical implication is clearest for REC-type trunk dictionaries in operator surrogates that predict wall-normal temperature or concentration profiles with a smooth outer part and a near-wall layer.

This advantage is expected to be particularly relevant to high-P\'eclet entrance-region heat and mass transfer problems in which repeated evaluations are required as inlet profiles or wall conditions change. For example, surface-based biosensors and biosensors based on surface capture require concentration profiles governed by convection, diffusion, reaction, and binding near reactive surfaces \cite{squires2008making}. Similarly, microfluidic systems with surface reactions and microfluidic electrochemical chips with electrodes on the side walls involve concentration fields formed by laminar convective-diffusive transport and electrochemical reaction at an electrode interface \cite{gervais2006mass,chevalier2021semianalytical}. In high-speed flow applications, high-speed boundary layers with wall transpiration, transpiration cooling through a porous wall, hypersonic transitional and turbulent boundary layers, hypersonic turbulent boundary layers with finite-rate chemical reactions, and reacting boundary layers with recombination reactions also require accurate near-wall temperature or species profiles \cite{sescu2019transpiration,hillcoat2025transpiration,xu2022hypersonic,passiatore2021finite,perakis2021recombination}.

\section{Conclusion}

This study examined trunk-basis design in DeepONet surrogates for singularly perturbed and high-P\'eclet transport operators on bounded domains, with particular emphasis on wall-normal profile reconstruction in entrance-region heat and mass transfer. The proposed REC trunk dictionary combines a low-degree outer Chebyshev subdictionary with an inner rational subdictionary constructed before DeepONet training from a canonical exponentially decaying layer family. The key conclusions are as follows:
\begin{enumerate}
    \item On the singularly perturbed scalar BVP, the \textit{REC} model gives its clearest reductions in the smallest-parameter regime, while remaining comparable to the \textit{Chebyshev} model over the full test set.

    \item On the thermal entrance problem, the \textit{REC} model improves the reconstruction of the wall-normal temperature profile $\theta(y)$ relative to both the \textit{Vanilla} and \textit{Chebyshev} models across the tested high-P\'eclet regime and sampled entrance locations $x$.

    \item On the concentration entrance problem, the \textit{REC} model improves the reconstruction of the wall-normal concentration profile $c(y)$ across $\mathrm{Da}=0.1$, $1.0$, and $10.0$, showing that the improvement persists as the absorbing-wall condition changes the near-wall layer.

    \item The reductions obtained with the \textit{REC} model are consistent across the five independent training runs and the held-out test profiles, indicating that the observed improvements are not restricted to a single training initialization or a small set of selected profiles.
\end{enumerate}

\section*{Acknowledgments}

This work was supported by the Center for Heterogeneous Integration of Micro Electronic Systems (CHIMES), one of the seven centers sponsored by the Semiconductor Research Corporation (SRC) and the Defense Advanced Research Projects Agency (DARPA) under the Joint University Microelectronics Program 2.0 (JUMP 2.0).

\section*{Declaration of competing interest}

The authors declare that they have no known competing financial interests or personal relationships that could have appeared to influence the work presented in this paper.

\section*{Data availability}

The code and data used in this work are available upon request.

\appendix

\section{Theoretical motivation for the REC trunk dictionary}\label{secA1}

This appendix develops an idealized representation argument for the REC trunk, rather than a convergence theorem for the trained DeepONet. The argument uses the canonical exponential layer family introduced in Eq.~\eqref{eq16}, whose sampled instances define the inner rational subdictionary in the REC dictionary construction before training. Treating the same family as an idealized target clarifies how polynomial and rational trunks represent a thin boundary-attached layer, and why the REC approximation space combining a low-degree outer Chebyshev subdictionary with the inner rational subdictionary is structurally compatible with the decomposition into a smooth outer solution and a one-sided layer. Define
\begin{equation}
s=2\zeta-1 \in [-1,1],
\qquad
\beta=\frac{1}{2\delta},
\end{equation}
where $s$ denotes the shifted coordinate and $\beta$ denotes the rescaled layer parameter. Then
\begin{equation}
\psi_{\delta}(\zeta)=e^{-\beta}e^{\beta s}.
\end{equation}
Using the modified-Bessel Jacobi-Anger identity \cite{dlmf1035},
\begin{equation}
e^{\beta s}=I_0(\beta)+2\sum_{k=1}^{\infty} I_k(\beta)T_k(s),
\end{equation}
where $I_k$ denotes the modified Bessel function of the first kind of order $k$ and $T_k$ denotes the Chebyshev polynomial of the first kind of degree $k$, one obtains the exact Chebyshev expansion
\begin{equation}
\psi_{\delta}(\zeta)
=
e^{-\beta}I_0(\beta)
+
2e^{-\beta}\sum_{k=1}^{\infty} I_k(\beta)\,T_k(2\zeta-1).
\label{eqa_chebyshev_expansion}
\end{equation}
Hence the Chebyshev coefficients are
\begin{equation}
c_0(\delta)=e^{-\beta}I_0(\beta),
\qquad
c_k(\delta)=2e^{-\beta}I_k(\beta), \quad k\ge 1,
\label{eqa_chebyshev_coefficients}
\end{equation}
where $c_k(\delta)$ denotes the coefficient of $T_k(2\zeta-1)$. For the degree-$n$ Chebyshev truncation
\begin{equation}
S_n\psi_{\delta}(\zeta)
=
c_0(\delta)+\sum_{k=1}^{n}c_k(\delta)\,T_k(2\zeta-1),
\end{equation}
where $n$ denotes the highest retained Chebyshev degree, the uniform truncation error satisfies
\begin{equation}
\|\psi_{\delta}-S_n\psi_{\delta}\|_{\infty}
\le
2e^{-\beta}\sum_{k=n+1}^{\infty} I_k(\beta),
\end{equation}
where $\|\cdot\|_{\infty}$ denotes the supremum norm on $[0,1]$. This shows that, as $\delta$ decreases, appreciable Chebyshev weight shifts to higher polynomial degrees. A low-degree polynomial trunk therefore becomes progressively less well matched to this canonical boundary layer family.

A more quantitative scale estimate follows from the local central-limit asymptotic for modified Bessel functions \cite{athreya1987bessel},
\begin{equation}
e^{-\beta}I_k(\beta)
\sim
\frac{1}{\sqrt{2\pi\beta}}
\exp\!\left(-\frac{k^2}{2\beta}\right),
\qquad
\beta\to\infty,
\qquad
\frac{k}{\sqrt{\beta}}\to\gamma\in[0,\infty),
\end{equation}
where $\gamma$ denotes the limiting scaled polynomial degree. This identifies $k=O(\sqrt{\beta})$ as the central coefficient scale. Since $\beta=1/(2\delta)$, this gives, for $k\ge 1$,
\begin{equation}
c_k(\delta)
\sim
2\sqrt{\frac{\delta}{\pi}}
\exp(-k^2\delta),
\qquad
c_0(\delta)
\sim
\sqrt{\frac{\delta}{\pi}}.
\label{eqa_chebyshev_coefficient_asymptotic}
\end{equation}
Thus, the envelope of the Chebyshev coefficients has a Gaussian scale in $k$. If $\rho\in(0,1)$ denotes a fixed relative decay level of this envelope and $K_{\rho}(\delta)$ denotes the polynomial degree at which $\exp(-k^2\delta)=\rho$, then
\begin{equation}
K_{\rho}(\delta)
=
\sqrt{\frac{|\log \rho|}{\delta}}.
\label{eqa_chebyshev_degree_scale}
\end{equation}
Therefore, the active polynomial degree scale is proportional to $\delta^{-1/2}$ as $\delta\to0$. For example, taking $\rho=e^{-1}$ gives $K_{\rho}(10^{-4})=100$, which explains why a degree $128$ Chebyshev trunk can remain a strong baseline for the pure canonical exponential layer. The inner rational subdictionary is therefore not motivated by a claim that it must dominate a sufficiently high degree Chebyshev expansion on this scalar layer alone. Rather, it is motivated as a compact layer-aligned inner rational subdictionary that can complement a low-degree outer Chebyshev subdictionary in the profile-valued entrance transport problems studied in the main text.

Next define the logarithmic parameter
\begin{equation}
\theta=\log_{10}\delta,
\end{equation}
where $\theta$ denotes the log scaled prototype parameter. Then, with $\delta=10^{\theta}$,
\begin{equation}
\partial_{\theta}\psi_{10^{\theta}}(\zeta)
=
(\ln 10)\left(\frac{1-\zeta}{\delta}\right)
\exp\!\left(-\frac{1-\zeta}{\delta}\right).
\end{equation}
If
\begin{equation}
t=\frac{1-\zeta}{\delta}\ge 0,
\end{equation}
where $t$ denotes the stretched layer coordinate associated with the canonical exponential profile, then
\begin{equation}
\partial_{\theta}\psi_{10^{\theta}}(\zeta)=(\ln 10)\,t e^{-t}.
\end{equation}
Since $t e^{-t}\le e^{-1}$ for all $t\ge 0$,
\begin{equation}
\sup_{\theta}\left\|\partial_{\theta}\psi_{10^{\theta}}\right\|_{\infty}
\le
\frac{\ln 10}{e},
\end{equation}
where the supremum is taken over the log parameter range considered for the layer family.

Let $\{r_i\}_{i=1}^{M}$ denote the rational dictionary elements, and let $\delta_i=10^{\theta_i}$ denote the prototype parameter value associated with $r_i$. Suppose that
\begin{equation}
\|\psi_{\delta_i}-r_i\|_{\infty}\le \tau_i,
\end{equation}
where $\tau_i$ denotes a uniform approximation error bound for the rational dictionary element $r_i$ on $[0,1]$. It should not be interpreted as the raw AAA stopping tolerance unless that tolerance has been separately verified to bound the uniform approximation error.

Then, for any $\delta=10^{\theta}$,
\begin{equation}
\|\psi_{\delta}-r_i\|_{\infty}
\le
\tau_i+\frac{\ln 10}{e}\,|\theta-\theta_i|
=
\tau_i+\frac{\ln 10}{e}\left|\log_{10}\delta-\log_{10}\delta_i\right|.
\end{equation}
Therefore,
\begin{equation}
\inf_{v\in\mathcal R_M}\|\psi_{\delta}-v\|_{\infty}
\le
\min_{1\le i\le M}
\left(
\tau_i+\frac{\ln 10}{e}\left|\log_{10}\delta-\log_{10}\delta_i\right|
\right),
\end{equation}
where
\begin{equation}
\mathcal R_M=\mathrm{span}\{r_1,\dots,r_M\}.
\end{equation}
Here, $\mathcal R_M$ denotes the rational subspace, $M$ denotes the number of rational dictionary elements, and $v$ denotes a candidate approximating function in $\mathcal R_M$.

Consequently, if $\mathcal I_{\theta}=[\theta_{\min},\theta_{\max}]$ denotes the sampled log parameter interval, and if
\begin{equation}
h_{\theta}
=
\sup_{\theta\in\mathcal I_{\theta}}
\min_{1\le i\le M}|\theta-\theta_i|
\end{equation}
denotes the fill distance of the sampled log grid, then
\begin{equation}
\inf_{v\in\mathcal R_M}\|\psi_{10^{\theta}}-v\|_{\infty}
\le
\tau_{\max}
+
\frac{\ln 10}{e}h_{\theta},
\qquad
\theta\in\mathcal I_{\theta},
\end{equation}
where $\tau_{\max}=\max_{1\le i\le M}\tau_i$ denotes the largest assumed uniform approximation error bound over the inner rational subdictionary. This is a log parameter coverage statement for the idealized layer family, not a statement about the trained DeepONet optimization error.

Finally, define
\begin{equation}
\mathcal P_{m-1}
=
\mathrm{span}\{T_0(2\zeta-1),\dots,T_{m-1}(2\zeta-1)\},
\qquad
\mathcal H_{m,M}=\mathcal P_{m-1}+\mathcal R_M,
\end{equation}
where $m$ denotes the number of outer Chebyshev dictionary elements and $\mathcal H_{m,M}$ denotes the corresponding REC approximation space.

For the \textit{REC} model used in the main comparison, $m=16$ and $M=113$. The Chebyshev trunk with the same nominal dimension corresponds to $\mathcal P_{128}$, which is obtained by taking $m=129$ in the definition of $\mathcal P_{m-1}$. In general, $\mathcal H_{16,113}$ and $\mathcal P_{128}$ are different approximation spaces, and neither space comparison alone implies that one trained DeepONet must outperform the other. The following estimate is therefore an existence bound for the REC approximation space, not a dominance theorem relative to the full Chebyshev space.

For a target field of the form
\begin{equation}
f(\zeta)=q(\zeta)+A\psi_{\delta}(\zeta),
\label{eqa18}
\end{equation}
where $f$ denotes a target profile, $q$ denotes a smooth outer component, and $A$ denotes a layer amplitude, let
\begin{equation}
\eta_m(q):=\inf_{\pi\in\mathcal P_{m-1}}\|q-\pi\|_{\infty},
\end{equation}
where $\pi$ denotes a candidate polynomial in $\mathcal P_{m-1}$, and $\eta_m(q)$ denotes the best uniform approximation error of the outer component $q$ by the outer Chebyshev subdictionary. Then
\begin{equation}
\inf_{v\in\mathcal H_{m,M}}\|f-v\|_{\infty}
\le
\eta_m(q)+|A|
\min_{1\le i\le M}
\left(
\tau_i+\frac{\ln 10}{e}\left|\log_{10}\delta-\log_{10}\delta_i\right|
\right).
\end{equation}
Although this bound does not prove that the trained \textit{REC} model must produce lower errors than the trained \textit{Vanilla} or \textit{Chebyshev} models in the comparisons of Section~\ref{sec3}, it shows that, under the idealized decomposition $f=q+A\psi_{\delta}$, the REC trunk is structurally aligned with a smooth outer component and a thin boundary-attached layer, while the inner rational subdictionary provides an error estimate controlled by the spacing of the sampled prototype parameter nodes on the $\log_{10}\delta$ axis for the canonical layer family.

The calculation also clarifies the interpretation of the singularly perturbed scalar BVP comparison in Section~\ref{sec3a}, where the \textit{REC} model improves mainly in the first three parameter bins but does not reduce every full-test metric relative to the \textit{Chebyshev} model. For the canonical exponential layer, the active polynomial degree scale is proportional to $\delta^{-1/2}$, indicating that a degree $128$ Chebyshev trunk remains an effective representation for $\delta\ge 10^{-4}$. Thus, the main role of the AAA-constructed inner rational subdictionary is not to replace high-degree Chebyshev approximation in this ideal scalar case, but to supply an explicitly layer-aligned component within a trunk dictionary having the same total number of dictionary elements when the target profiles combine smooth outer behavior with wall-attached transport layers.

\end{document}